\documentclass{article}

\usepackage{microtype}
\usepackage{graphicx}
\usepackage{booktabs} %

\usepackage{multirow}
\usepackage{graphicx}
\usepackage{xcolor,colortbl}

\usepackage{hyperref}

\usepackage[accepted]{icml2023}
\newcounter{daggerfootnote}

\usepackage{caption}
\usepackage{subcaption}
\DeclareCaptionFormat{custom}{\textsl{#1#2}#3}
\usepackage[flushleft]{threeparttable}

\usepackage{amsmath}
\usepackage{amssymb}
\usepackage{mathtools}
\usepackage{amsthm}

\usepackage[capitalize,noabbrev]{cleveref}

\theoremstyle{plain}

\theoremstyle{definition}

\theoremstyle{remark}

\usepackage{inconsolata}

\graphicspath{{attn_maps}}

\usepackage{comment}
\usepackage[detect-none]{siunitx}
\usepackage{tikz}
\usepackage{pgfplots}
\pgfplotsset{compat=1.16}
\usetikzlibrary{fit,arrows.meta,mindmap,shadows,backgrounds,calc,positioning,shapes.misc}

\let\cite\citep

\newcommand{\fairseq}{\textsc{fairseq}\xspace}

\newcolumntype{a}{>{\columncolor{blue!5}}c}

\definecolor{cadetgrey}{rgb}{0.57, 0.64, 0.69}

\input{math}

\DeclareMathOperator{\enc}{\textsc{encode}}
\DeclareMathOperator{\collapse}{\textsc{collapse}}
\DeclareMathOperator{\bce}{\textsc{bce}}
\newcommand{\blank}{\textvisiblespace}
\newcommand{\sub}[1]{_{\mathrm{#1}}}

\usepackage{xspace}
\newcommand{\covost}{\mbox{CoVoST}\xspace}
\newcommand{\covostt}{\mbox{CoVoST-2}\xspace}
\newcommand{\mustc}{\mbox{MuST-C}\xspace}

\usepackage[textsize=small]{todonotes}
\setuptodonotes{backgroundcolor=white, bordercolor=white}
\usepackage{marginnote}

\definecolor{carolinablue}{rgb}{0.6, 0.73, 0.89}
\newcommand{\perate}[1]{\textcolor{carolinablue}{(#1)}}

\makeatletter
\renewcommand{\paragraph}{%
	\@startsection{paragraph}{4}{\z@}{0.5ex plus
   0.5ex minus .2ex}{-1em}{\normalsize\bf}%
}
\makeatother

\title{Pre-training for Speech Translation: CTC Meets Optimal Transport}

\icmltitlerunning{Pre-training for Speech Translation: CTC Meets Optimal Transport}

\begin{document}

\twocolumn[
\icmltitle{Pre-training for Speech Translation: CTC Meets Optimal Transport}

\icmlsetsymbol{equal}{*}

\begin{icmlauthorlist}
\icmlauthor{Phuong-Hang Le}{uga}
\icmlauthor{Hongyu Gong}{meta}
\icmlauthor{Changhan Wang}{meta}
\icmlauthor{Juan Pino}{meta}
\icmlauthor{Benjamin Lecouteux}{uga}
\icmlauthor{Didier Schwab}{uga}
\end{icmlauthorlist}

\icmlaffiliation{uga}{Univ.~Grenoble Alpes, CNRS, Grenoble INP, LIG, 38000 Grenoble, France}%
\icmlaffiliation{meta}{Meta AI}

\icmlcorrespondingauthor{Phuong-Hang Le}{hang.le@univ-grenoble-alpes.fr}

\icmlkeywords{Machine Learning, ICML}

\vskip 0.3in
]

\printAffiliationsAndNotice{}  %

\begin{abstract}
The gap between speech and text modalities is a major challenge in speech-to-text translation~(ST). Different methods have been proposed to reduce this gap, but most of them require architectural changes in ST training. In this work, we propose to mitigate this issue at the pre-training stage, requiring no change in the ST model. First, we show that the connectionist temporal classification (CTC) loss can reduce the modality gap by design. We provide a quantitative comparison with the more common cross-entropy loss, showing that pre-training with CTC consistently achieves better final ST accuracy. Nevertheless, CTC is only a partial solution and thus, in our second contribution, we propose a novel pre-training method combining CTC and optimal transport to further reduce this gap. Our method pre-trains a Siamese-like model composed of two encoders, one for acoustic inputs and the other for textual inputs, such that they produce representations that are close to each other in the Wasserstein space. Extensive experiments on the standard \covostt and \mustc datasets show that our pre-training method applied to the \emph{vanilla} encoder-decoder Transformer achieves state-of-the-art performance under the no-external-data setting, and performs on par with recent strong multi-task learning systems trained with external data. Finally, our method can also be applied on top of these multi-task systems, leading to further improvements for these models.%
\end{abstract}

\section{Introduction}\label{sec:introduction}%

Speech-to-text translation~(ST) is a challenging task that often requires training two auxiliary tasks for better performance: automatic speech recognition~(ASR) and machine translation~(MT). This can be achieved through either \emph{pre-training}~\cite{berard2018end,bansal2019pretraining,wang2020curriculum,wang2020fairseq,inaguma2020espnet,le2021lightweight} 
or \emph{multi-task learning} (joint-training) approaches~\cite{anastasopoulos2018tied,sperber2019attention,le2020dual,chuang2020worse,wang2020bridging,tang2021general,tang2021improving,ye2022cross}. In particular, ASR and MT pre-trainings have become arguably the most standard practice in ST development. Indeed, they have been used to obtain strong baselines in popular libraries such as \mbox{ESPnet-ST}~\cite{inaguma2020espnet} and \mbox{\fairseq-S2T}~\cite{wang2020fairseq}, or in standard benchmarks such as \mustc~\cite{di2019must} and \covost~\cite{wang2020covost,wang2020covost2}. Furthermore, they have also been adopted in most of the submissions to recent IWSLT evaluation campaigns~\cite{anastasopoulos2021findings,anastasopoulos2022findings}, %
as well as in strong multi-task learning systems~\cite{tang2021general,tang2021improving,tang2022unified}. Despite such ubiquity, however, this approach presents a major limitation, namely the so-called \emph{modality gap}. Indeed, the ASR encoder is pre-trained with speech inputs, whereas the MT decoder is pre-trained with text inputs, %
thus plugging them together (for ST fine-tuning) will naturally result in a mismatch. This explains why simply using a pre-trained MT decoder in addition to a pre-trained ASR encoder only brings modest gains~\cite{alinejad2020effectively} or sometimes even worsens the performance~\cite{bahar2019comparative}. 

In this work, we make two major contributions for mitigating the modality gap without requiring any changes in the ST model. First, we show that CTC~\cite{graves2006connectionist} is a viable solution to this problem. We advocate through theoretical analysis and extensive experiments the use of CTC for ASR pre-training over the standard cross-entropy~(CE) loss. Indeed, pre-training with \emph{vanilla} CTC (\ie, without additional components such as an external language model) is not only faster but also yields better final ST results than pre-training with CE. At first glance, this is rather counter-intuitive, given that vanilla CTC generally produces inferior ASR performance than CE (as already noticed by previous work~\cite{bahdanau2016end2end,kim2017jointCTC} and again confirmed by our experiments); but as we will point out, this success of CTC may be attributed to the ability of its trained encoder to align speech input to text output \emph{without the need for a decoder}.
Even so, however, CTC can only partially alleviate the modality gap. Therefore, in our second contribution, we propose a novel pre-training method combining CTC and optimal transport (OT)~\cite{peyre2019computational} to further reduce this gap. It consists in training a Siamese-like model composed of two encoders, one for acoustic inputs and the other for textual inputs, such that they produce representations (for the same sentence)  %
close to each other in the Wasserstein space, \ie, a metric space equipped with the Wasserstein distance from OT. Furthermore, we introduce \emph{positional encoding} to the given OT formulation in order to take into account the \emph{monotonicity} of the inputs, which brings an extra boost in performance. The proposed method can work with or without MT pre-training, and in the former case, it can use the pre-trained MT decoder in a more effective way. Extensive experiments on the standard \covostt~\cite{wang2020covost2} and \mustc~\cite{di2019must} datasets show that our pre-training method applied to the \emph{vanilla} encoder-decoder Transformer achieves  state-of-the-art performance under the no-external-data setting. Moreover, simply increasing the model size and using our method, still without using any additional data, leads to performance that is competitive with recent strong multi-task learning systems trained with external data. Finally, our pre-training method can also be applied on top of these multi-task systems, leading to further improvements. Our code is available at 
\href{https://github.com/formiel/fairseq}{\texttt{github.com/formiel/fairseq}}.
\vspace{-2pt}

\section{Related Work}\label{sec:related-work}

\paragraph{Speech-to-text translation} The classical approach to ST consists in using \emph{cascaded} systems composed of an ASR module followed by an MT one~\cite{ney1999speech}. This approach, however, presents important limitations such as having high latency and being susceptible to error propagation~\cite{anastasopoulos2018tied}. Recently, much effort has been put into exploring \emph{end-to-end} ST models~\cite{duong2016attentional,berard-nips2016,weiss2017sequence,di2019must,inaguma2019multilingual,le2020dual}. 
While the first works in this direction 
only obtained modest results~\cite{berard-nips2016,weiss2017sequence}, the most recent ones have largely closed the gap with cascaded models, or even surpassed them~\cite{bentivogli2021cascade,xu2021stacked,ye2021end,tang2021improving,ye2022cross}. Our work falls into the end-to-end paradigm as a generic pre-training approach that can be applied on top of existing methods.\vspace{-3pt}

\paragraph{Reducing pre-training modality gap} Various methods have been studied to bridge the gap between pre-training and fine-tuning such as using cascaded encoders~\cite{liu2020bridging,wang2020bridging,xu2021stacked}, adaptor modules~\cite{bahar2019comparative,li2020multilingual,xu2021stacked}, or adversarial regularizer~\cite{alinejad2020effectively}. Multi-task learning~\cite{tang2021general,tang2021improving,tang2022unified} can also be seen as a solution to this problem. However, most of these methods do not tackle the issue at the pre-training stage but adapt the ST architecture to suit 
the pre-trained weights. Our method, instead, can mitigate the modality gap at the pre-training stage without requiring any changes in the ST model.

\paragraph{CTC for ST pre-training} CTC has been used for pre-training in previous work, either alone~\cite{wang2020bridging} or in combination with cross-entropy~\cite{zhang2020adaptive,wang2020curriculum,xu2021stacked,inaguma2021source} or other losses~\cite{bapna2022mslam}. However, no detailed quantitative comparison with cross-entropy was given and thus it was not clear how much CTC actually contributed to the obtained improvements in these works. Our work can be seen as complementary to them in terms of quantitative analysis of CTC pre-training. We believe that this is an important contribution as the effectiveness of CTC pre-training has gone relatively unnoticed by the community. For example, cross-entropy has still been used by default in the most recent multi-task learning systems~\cite{tang2021general,tang2021improving,tang2022unified}. We show that simply replacing it with CTC leads to improvements for these methods.

\paragraph{Aligning speech and text}Learning to align speech and text features has been considered previously for ST, \eg, in supervised pre-training (similar to our setting) using an adversarial loss~\cite{alinejad2020effectively}, in self-supervised pre-training~\cite{bapna2021slam,bapna2022mslam,chen2022maestro,ao2022speecht5,ao2022pretraining}, and in multi-task learning using Euclidean distance~\cite{liu2020bridging,dong2021listen,tang2021improving}, cosine distance~\cite{chuang2020worse}, Kullback–Leibler divergence~\cite{tang2022unified}, and contrastive loss~\cite{han2021learning,ye2022cross,ouyang2022waco}. We should point out that our architecture is conceptually simpler than those proposed in these works. Moreover, our framework can also work with OT replaced by any (differentiable) distance function, including those aforementioned. A major strength of OT, in addition to its strong theoretical properties and practical performance, is that it can \emph{natively} handle sequences of different lengths, whereas Euclidean distance and Kullback–Leibler divergence require some length-matching mechanism such as average pooling~\cite{liu2020bridging,dong2021listen} or attention~\cite{tang2021improving}. Finally, we should note that speech-text alignment has been used before in other speech applications beyond ST, such as ASR~\cite{juang1984hidden,berndt1994using}, text-to-speech~\cite{haubold2007alignment}, or speaker identification~\cite{yuan2008speaker,muda2010voice}, with many of them employing dynamic time warping~(DTW)~\cite{sakoe1978dynamic}. A more complete survey on this topic can be found in~\citet[Section 4]{liang2022foundations}.

\paragraph{OT for speech translation} OT has widely been used in MT~\cite{alqahtani2021using,alvarez2019towards,alvarez2018Gromov-Wasserstein,grave2019unsupervised,chen2019improving}
and also in speech processing, \eg, for speech enhancement~\cite{lin2021unsupervised}. OT has not been used for speech translation, to the best of our knowledge.

\section{The Modality Gap in ST Pre-training}\label{sec:modality-gap}

\begin{figure}[!htb]
\begin{subfigure}[b]{\linewidth}\centering%
\includegraphics[width=\linewidth]{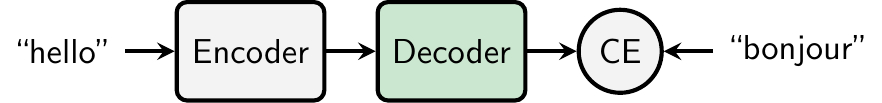}%
\caption{\label{fig:mt-pre-training}\textbf{MT pre-training (optional)} Standard encoder-decoder with cross-entropy (CE) loss. The encoder is often discarded for ST.}
\end{subfigure}\\%

\begin{subfigure}[b]{\linewidth}\centering%
\includegraphics[width=\linewidth]{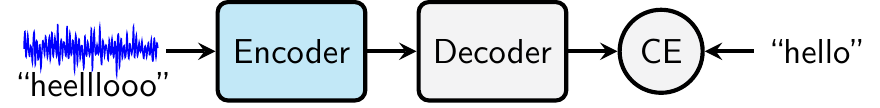}%
	\caption{\label{fig:asr-pre-training}\textbf{ASR pre-training} Standard encoder-decoder with CE loss. The decoder is often discarded for ST fine-tuning.}
\end{subfigure}	\\

\begin{subfigure}[b]{\linewidth}\centering%
\includegraphics[width=\linewidth]{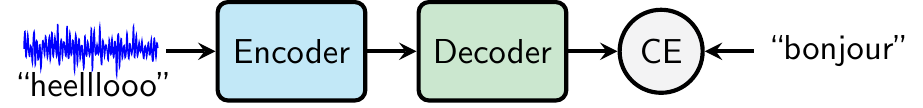}%
\caption{\label{fig:st}\textbf{ST fine-tuning} Using ASR encoder and MT decoder (if available) for initialization.}
\end{subfigure}
	\caption{\label{fig:recipe-basic}Standard pre-training recipe for ST.}
\end{figure}

We review the most standard pre-training recipe for ST (namely ASR and MT pre-training, Figure~\ref{fig:recipe-basic}) and discuss the induced gap between pre-training and fine-tuning. %
In a few words, we pre-train ASR and MT using the standard encoder-decoder architecture with cross-entropy loss, then the ASR encoder and MT decoder are used to initialize the corresponding ST components. Note that MT pre-training is optional as it does not necessarily improve the performance, as discussed in Section~\ref{sec:introduction}. Here we assume that the ST model is also the vanilla encoder-decoder. In more sophisticated models such as multi-task learning~\cite{tang2021improving,ye2022cross}, all pre-trained components are typically used.

In an attention-based encoder-decoder model, the decoder typically learns to \emph{align} the output with the input~\cite{bahdanau2015neural}. In the above recipe, the two components of ST are pre-trained to be aligned with other components that are later discarded, causing a \emph{loss of alignment information}. This explains the modality discrepancy between pre-training and fine-tuning. In the next section, we provide an explanation why CTC can partially solve this problem.

\section{ASR Pre-training with CTC}
\label{sec:CTC-pre-training}

In this section, we revisit CTC~\cite{graves2006connectionist} for ASR pre-training and explain why it can be seen as a partial solution to the modality gap issue.

\subsection{Review of CTC}\label{sec:ctc-review}

Given an input sequence of (typically pre-processed and downsampled) audio features $\X\triangleq (\x_1,\dots,\x_{S})$ (in some language, \eg, English), the task of ASR consists in predicting its transcription (in the same language), represented by an output sequence $\y \triangleq (y_1,\dots,y_{T})$, where $y_t\in \cV \triangleq \set{1,2,\dots, V}$, a vocabulary of size $V$. Suppose that $\X$ is processed by an encoder to obtain
\begin{equation}
    \H \triangleq (\h_1,\dots,\h_S) \triangleq \enc(\X;\thetab\sub{enc}),
\end{equation}
where $\h_t\in\RR^d$ is the hidden feature vector at time step $t\in\set{1,\dots,S}$, and $\thetab\sub{enc}$ is the parameters of the encoder. The idea of CTC is to predict a token $\hat{a}_t\in\cV$ for each time step $t$ based on $\h_t$:
\begin{align}
	p(a_t\given \X) &= \softmax(\W\h_t + \b)[a_t] \ \forall a_t\in\cV,\nonumber\\
	\hat{a}_t &= \argmax_{a_t\in\cV} p(a_t\given \X),\label{eq:ctc-alignment}
\end{align}
where $\W\in\RR^{V\times d}, \b\in\RR^{V}$ are the weights and biases of the final linear layer, and $\v[i]$ denotes the $i\textsuperscript{th}$ element of a vector $\v$. The vector $\hat{\a}\triangleq (\hat{a}_1,\dots,\hat{a}_S)$ is called an \emph{alignment}. Clearly, this produces a sequence of tokens of the same length as the input, which is not desirable. CTC solves this by applying a collapsing function, $\y = \collapse(\hat{\a})$, that removes consecutive repetitions. It is assumed that the vocabulary contains a special \emph{blank} token (``\blank'', which the model can predict), and collapsing only happens in-between blanks and not across them:
\begin{align*}
	\collapse(\texttt{heellllloooo}) &= \texttt{helo}\\
	\collapse(\texttt{he\blank ll\blank l\blank oo\blank\blank}) &= \texttt{hello}.
\end{align*}
The issue is that different $\hat{\a}$ may collapse to the same $\y$, and thus the most likely assignment (given by~\eqref{eq:ctc-alignment}) may not correspond to the most probable final output. To obtain the latter, CTC actually computes the highest sum over the probability of all its possible alignments using the Viterbi algorithm~\cite{viterbi1967error}. We refer
the reader to~\citet{graves2006connectionist} for further details.

\subsection{CTC can reduce modality gap in pre-training}\label{sec:ctc_reduces_gap}

\begin{figure}[!htb]
\includegraphics[width=\linewidth]{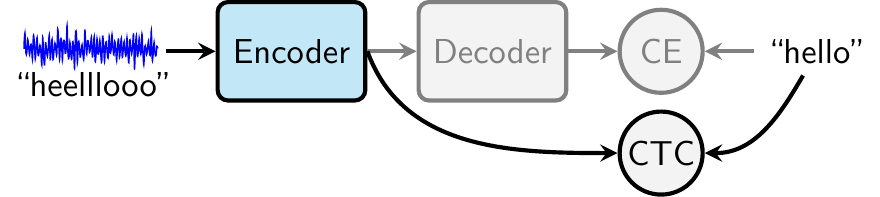}%
	\caption{\label{fig:ctc-pre-training}Two flavors of ASR pre-training with CTC: with and without CE. In the latter, the CE branch (in gray) is not present.}
\end{figure}

CTC has been used for pre-training in the ST literature, either alone~\cite{wang2020bridging} or with cross-entropy (CE)~\cite{zhang2020adaptive,wang2020curriculum,xu2021stacked,inaguma2021source}. This is illustrated in Figure~\ref{fig:ctc-pre-training} (compare with Figure~\ref{fig:asr-pre-training}).

\noindent We will see later in the experiments that CTC completely outperforms CE in terms of ST accuracy, while being on par with CE+CTC, which indicates its importance for pre-training. There is a simple explanation for this. On one hand, recall from our discussion in Section~\ref{sec:modality-gap} that the discrepancy between pre-training and ST fine-tuning is caused by the loss
of alignment information after discarding the ASR decoder and the MT encoder. On the other hand, recall from Section~\ref{sec:ctc-review} that \textbf{the ASR encoder trained with CTC already learns to align speech input to text output} without a decoder, which means that \emph{no alignment information will be lost} when the encoder is used for ST fine-tuning, as the encoder already contains this information. That is, even though the model trained with CE is stronger (than CTC) in general~\cite{bahdanau2016end2end,kim2017jointCTC}, the contribution of the CE decoder to the overall performance could be so important that removing it would make the CE encoder (alone) weaker than the CTC encoder. This is indeed confirmed by our experiments (Appendix~\ref{sec:appendix-asr_performance_speech_encoders}).

Finally, CTC obviously cannot recover the loss
of alignment information when discarding the MT encoder, and thus it should be considered only as a partial solution. In the next section, we present a method for filling this gap.

\section{Optimal Transport for Pre-training }\label{sec:ot}

This section presents our core contribution for reducing the modality gap in ST. The idea is to train the speech encoder to generate representations that are close to those produced by a text encoder. The challenge here is that, given the same sentence, its speech features typically have a \emph{much longer sequence length} than its text features, which makes it difficult to ``compare'' them. The Wasserstein distance from optimal transport (OT) turns out to be a suitable solution. %

\subsection{Review of discrete optimal transport}\label{sec:ot-review}

We first give a brief review of OT~\cite{peyre2019computational}. 

Let $\alpha$ be a discrete probability distribution represented by positive masses $a_1,\dots,a_m$ (where $a_1+\dots+a_m=1$) at locations $\u_1,\dots,\u_m\in\RR^d$, respectively (\ie, $a_i$ is the quantity of mass at $\u_i$). Suppose we want to form a new distribution $\beta$ with masses $b_1,\dots,b_n$ (where $b_1+\dots+b_n=1$) at new locations $\v_1,\dots,\v_n\in\RR^d$ by transporting all the masses from $\alpha$ to $\beta$. Suppose that the \emph{cost} of transporting a unit of mass from $\u_i$ to $\v_j$ is given by $c(\u_i,\v_j)$, where $c:\RR^d\times\RR^d \to \RR_+$ is some function.
Let $Z_{ij} \ge 0$ be the quantity of mass to be transported from $\u_i$ to $\v_j$, which induces a cost of $Z_{ij}c(\u_i,\v_j)$. The total transportation cost is thus $\sum_{i=1}^m\sum_{j=1}^n Z_{ij}c(\u_i,\v_j)$.
The total quantity of mass that $\beta$ receives from $\u_i$ is $\sum_{j=1}^n Z_{ij}$, which must equal the mass stored at $\u_i$, thus $\sum_{j=1}^n Z_{ij} = a_i$. Similarly, the total quantity of mass that $\v_j$ receives from $\alpha$ is $\sum_{i=1}^m Z_{ij}$, and thus $\sum_{i=1}^m Z_{ij} = b_j$. OT consists in finding the \emph{transportation plan} $\Z^*$ that has the minimum cost:%
\begingroup
\setlength\abovedisplayskip{10pt}
\setlength\belowdisplayskip{5pt}
\begin{equation}\label{eq:ot}
    \min_{\Z}\inner{\C, \Z} \
	\mbox{s.t.} \ \Z\1_n = \a, \Z^\top\1_m = \b, \Z\ge\0,
\end{equation}
\endgroup
where $\1_n$ denotes the $n$-dimensional vector of ones, $\a = (a_1,\dots,a_m)$, $\b=(b_1,\dots,b_n)$, $\Z$ and $\C$ denote the $m\times n$ matrices whose elements are $Z_{ij}$ and $C_{ij}= c(\u_i,\v_j)$. %

\begin{figure*}[!htb]\centering%
\includegraphics[]{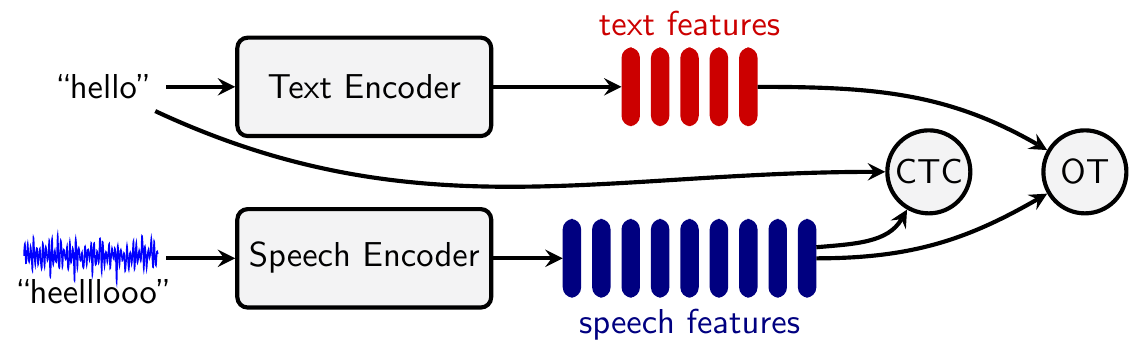}%
	\caption{\label{fig:speech-text-alignment}Our proposed Siamese-like architecture for learning to align speech and text representations. An input pair of audio sequence and its transcript are fed to the corresponding speech and text encoders, then their outputs are compared using optimal transport (OT). The speech features are enhanced by jointly training with CTC.}%
\end{figure*}

\paragraph{Wasserstein loss} Let $\Z^*$ denote the optimal solution to~\eqref{eq:ot}. If we define $W(\alpha,\beta) = \inner{\C, \Z^*}$ (\ie, the minimum transportation cost), then $W$ can be seen as a distance measure between $\alpha$ and $\beta$, and is called the \emph{Wasserstein distance}. %
In theory, one can use $W$ as a loss function (called the \emph{exact} Wasserstein loss~\cite{frogner2015learning}) because it is differentiable almost everywhere. However, evaluating this loss requires solving~\eqref{eq:ot}, which is expensive in practice. It is typically better to work with 
an upper-bound approximation of $W$, defined as (subject to the same constraints as in~\eqref{eq:ot})
\begin{equation}\label{eq:wasserstein-regularized}
    W_{\lambda}(\alpha,\beta) = \min_{\Z}\set{\inner{\C, \Z} - \lambda H(\Z)},
\end{equation} 
where $H$ is the entropy function and $\lambda > 0$ is a regularization weight. The function $W_\lambda$ is not only fully differentiable but also very efficient to evaluate using the so-called Sinkhorn algorithm~\cite{sinkhorn1967concerning}. We refer the reader
to~\citet{cuturi2013sinkhorn} and~\citet{frogner2015learning} for details. From now on, ``Wasserstein distance'' (or loss) refers to this regularized variant $W_\lambda$.

\subsection{Learning to align speech and text features}
\label{sec:learning-to-align}

In this section, we present our proposed model for learning to align speech and text features. Recall that our goal was to mitigate the modality gap issue arising in ST, which involves the discrepancy between acoustic and textual representations. To this end, we propose an architecture composed of two encoders, one for speech inputs and the other for text inputs.
Then, given an input pair of an audio sequence and its transcript, we feed them to the corresponding encoders and train the model to produce features that are close to each other in terms of Wasserstein distance. In addition, as ASR data (audio together with transcripts) are
assumed to be available for this model, we make further use of them
to enrich the learned speech representations by jointly training with a CTC loss. This is summarized in Figure~\ref{fig:speech-text-alignment}.

\paragraph{Wasserstein distance between speech and text} We have mentioned the Wasserstein distance between two sets of features, without having formally defined it (note that the previous section only presents the Wasserstein distance between two probability distributions). Let $\U = (\u_1,\dots,\u_m)$ and $\V = (\v_1,\dots,\v_n)$ be the output speech and text features, respectively, where $\u_i\in\RR^d$ and $\v_j\in\RR^d$ for all $i,j$. Here we have assumed that the hidden dimensions of the two encoders are the same and equal to $d$, but the sequence lengths $m$ and $n$ can be different (typically $m$ is much larger than $n$). Define two distributions $\alpha$ and $\beta$ whose masses are uniformly distributed at locations $(\u_1,\dots,\u_m)$ and $(\v_1,\dots,\v_n)$, respectively; here uniform distribution means that $a_i=\frac{1}{m}$ and $b_j=\frac{1}{n}$ for all $i,j$, where $\a$ and $\b$ are defined as in the previous section. Define the cost of transporting a unit of mass from $\u_i$ to $\v_j$ as $c(\u_i,\v_j) = \norm{\u_i-\v_j}_p$ for some $p\ge 1$ (typically $p=2$). Then, the Wasserstein distance $W_\lambda(\alpha,\beta)$ (see~\eqref{eq:wasserstein-regularized}) can be seen as the discrepancy between $\U$ and $\V$. Without ambiguity, we refer to this quantity as the Wassertein distance between these two sequences. The obtained optimal transportation plan can then be seen as an alignment map between the two sequences.

\paragraph{Positional encoding for OT} An important property of the above Wasserstein distance between two sets of features is that it does not take into account the sequence orders. Indeed, if we replace one of the sequences by any of its permutations, then the Wasserstein distance remains the same. While this property can be useful in many situations, taking into account the sequence orders could be beneficial in our case due to the nature of the task: the inputs to our encoders are \emph{monotonically} aligned. That is, the first audio frames should represent the same part of the sentence as the first text tokens do, and vice-versa (note that this is true for ASR but not for MT or ST). We propose to integrate this prior information into the OT model as follows. The idea is to modify the cost matrix $\C$ such that transporting from one location to another will have a high cost if their positions in the corresponding sequences are very different. For example, transporting from $\u_1$ to $\v_1$ (or from $\u_m$ to $\v_n$) should induce a low cost while transporting from $\u_1$ to $\v_n$ should induce a high cost. As the input sequence lengths can be very different, we normalize them to unit length so that the first element of the sequence has position $0$ and the last has position $1$. That is, the position vectors $(1,2,\dots,m)$ and $(1,2,\dots,n)$ will be normalized to, respectively, $(s_1,s_2,\dots,s_m)$ and $(t_1,t_2,\dots,t_n)$, where 
\begin{equation}
    s_i = \frac{i-1}{m-1}\quad \mbox{ and }\quad t_j = \frac{j-1}{n-1}.
\end{equation}
Then, a simple way of including the positional constraint into the cost matrix could be to define, \eg, $C_{ij} = c(\u_i,\v_j) + \gamma \abs{s_i-t_j}$ (where $\gamma > 0$ is a weight scalar that can be empirically tuned), which would penalize any mismatch in terms of positions.\footnote{It is worth pointing out that, even though the alignment of the speech and text features is supposed to be monotonic, it is not necessarily \emph{linear}. For example, the speech features at (normalized) position $0.5$ need not correspond to the text features at position $0.5$. Thus, our mismatch cost can be viewed as favoring a linear approximation of the true alignment (which is unknown).} However, this may cause inefficiency in practice. Indeed, in our case $c$ is an $\ell_p$-norm for some $p\ge 1$ (according to the previous paragraph), for which the Wasserstein distance can be evaluated very efficiently without computing and storing the matrix $\C$ explicitly~\cite{feydy2019interpolating}; the above modification of $\C$ could make $C_{ij}$ no longer an $\ell_p$-norm (unless $p=1$), thus losing this efficiency. Therefore, given that $c$ is an $\ell_p$-norm, we propose to modify the cost matrix as follows:
\begingroup
\setlength\abovedisplayskip{2pt}
\setlength\belowdisplayskip{0pt}
  \begin{equation}\label{eq:positional_ot}
    C_{ij} = \left(\norm{\u_i-\v_j}_p^p + \gamma^p \abs{s_i-t_j}^p\right)^{1/p}.
\end{equation}
\endgroup

Put 
$\u'_i = [\u_i; \gamma s_i]$ and $\v'_j = [\v_j; \gamma t_j]$, 
then the above is just $\|\u'_i - \v'_j\|_p$. That is, we have constructed a new feature vector at each sequence element by appending its normalized position to its existing feature vector, so that performing OT on these new features is equivalent to performing it on the original features using the new cost matrix defined by~\eqref{eq:positional_ot}. %

\paragraph{Beyond optimal transport} Clearly, the proposed method can also work with OT replaced by any differentiable loss function, including (soft) DTW~\cite{cuturi2017soft}, adversarial loss~\cite{ganin2016domain}, Euclidean distance, or KL divergence (the latter two cannot work directly on sequences of different lengths but require some length-matching mechanism such as average pooling~\cite{liu2020bridging,dong2021listen} or attention~\cite{tang2021improving}; see also Section~\ref{sec:related-work} and Appendix~\ref{sec:appendix-distance-functions} for detailed descriptions of these methods). We will see in later experiments that OT turns out to have the best practical performance (in addition to its strong theoretical properties, such as being a metric).

\begin{figure}[!t]
\begin{subfigure}[b]{\linewidth}\centering%
\includegraphics[width=\linewidth]{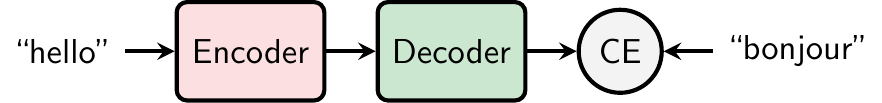}%
\caption{\label{fig:mt_ours}\textbf{MT pre-training} Standard and same as Figure~\ref{fig:mt-pre-training}.}
\end{subfigure}\\%

\begin{subfigure}[b]{\linewidth}\centering%
\includegraphics[width=.7\linewidth]{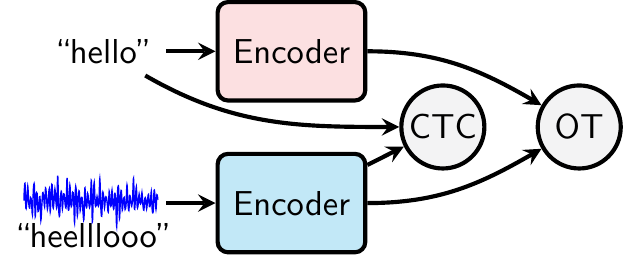}%
\caption{\label{fig:asr-full-ours}\textbf{ASR pre-training} Speech encoder and text encoder learn to produce similar representations in the Wasserstein space. The text encoder is initialized with the pre-trained MT encoder.}
\end{subfigure}\\%

\begin{subfigure}[b]{\linewidth}\centering%
\includegraphics[width=\linewidth]{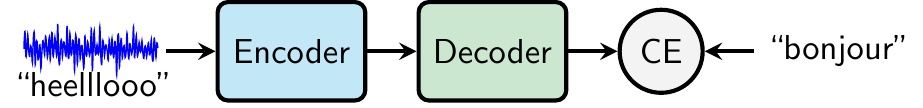}%
\caption{\label{fig:st-ours}\textbf{ST fine-tuning} Both pre-trained ASR encoder and MT decoder are used for initialization.}
\end{subfigure}
	\caption{\label{fig:recipe-ours}Our proposed pre-training recipe for ST.}\vspace{-10pt}
\end{figure}

\subsection{Proposed recipes for speech translation}
\label{sec:proposed-recipe}

Equipped with the above method for learning to align acoustic and textual representations, we are now ready to present our proposed training recipe for ST.
As shown in Figure~\ref{fig:recipe-ours}, our recipe differs from the standard one (Figure~\ref{fig:recipe-basic}) only at the ASR pre-training stage. We follow the above method (Section~\ref{sec:learning-to-align}) and train a Siamese-like model using CTC and OT. Here the text encoder is initialized with the pre-trained MT encoder, which helps the speech encoder learn to align with the text decoder right at the pre-training stage. %
Note that freezing the text encoder during ASR pre-training also produced good results in our experiments, though not as good as training both encoders. In addition, a simplified variant of our recipe can also be obtained by omitting MT pre-training. In that case, the text encoder in ASR pre-training is initialized randomly. As shown later in Section~\ref{sec:experiments}, this recipe also outperforms ASR pre-training with CTC and cross-entropy, indicating the effectiveness of OT.

\section{Experiments}\label{sec:experiments}
In this section, we provide empirical evidence for the effectiveness and versatility of our proposed method. Section~\ref{sec:experimental-setup} outlines the experimental setup, while Section~\ref{sec:ablation-analysis} conducts ablation studies to justify our various design choices. Comparative results of the presented pre-training methods are discussed in Section~\ref{sec:results-dev-set}, and comparison to state-of-the-art methods is presented in Section~\ref{sec:test_results}. Finally, Section~\ref{sec:multi-task} demonstrates that our pre-training method can also be applied to multi-task learning, showing its generality.

\subsection{Experimental setup}\label{sec:experimental-setup}

\paragraph{Datasets} We evaluate the pre-training methods presented in this paper on the standard
\textbf{\mustc}~\cite{di2019must} and \textbf{\covostt}~\cite{wang2020covost2} datasets. \mustc is a large-scale one-to-many ST dataset built from audio recordings of TED Talks, covering pairs from English~(En) to 8 European languages: Dutch~(Nl), French~(Fr), German~(De), Italian~(It), Portuguese~(Pt), Romanian~(Ro), Russian~(Ru), and Spanish~(Es). Each direction includes a triplet of speech, transcription, and translation. %
\covostt is a large and diversified multilingual ST corpus based on the Common Voice project~\cite{ardila2020common}. It covers translations from 21 source languages into English and from English into 15 target languages, many of which involves non-European and very low-resource languages. We refer the reader to the original papers for further details. 
It is important to note that all our analyses are conducted on the \emph{dev} splits of these datasets to prevent overfitting their \emph{test} sets. Then, only the best-performing models will be selected for comparison with existing methods on the \emph{test} sets.

\paragraph{Training settings} While we focus on \emph{multilingual} translation (\ie, a single model for all language pairs), our ablation analysis is conducted under the \emph{bilingual} setting with only two language pairs (one model for each), due to the high training cost. For pre-training, we consider two settings: \emph{with} and \emph{without} MT pre-training. In the latter, some components are initialized randomly (see Figures~\ref{fig:recipe-basic} and~\ref{fig:recipe-ours}).

\paragraph{Implementation details} Our implementation is based on the \textsc{fairseq S2T} toolkit~\cite{wang2020fairseq}. We follow closely previous work~\cite{wang2020fairseq,wang2020covost2} for setup, including models, data processing, training and evaluation settings (see Appendix~\ref{sec:implentation-details} for details). %
In particular, we use a model whose speech encoder, speech decoder, and text decoder have respectively $12$, $6$,
and $6$ layers. 
For the analysis, we use a \emph{medium} architecture with hidden dimension $d=512$. In the final experiments where we aim to reach state-of-the-art performance for comparison with existing methods, we also use the \emph{large} variant where $d=1024$.

\paragraph{Loss functions} As CTC plays a crucial role in the final ST performance, when training it jointly with another loss (CE or OT), we keep it unchanged and scale the latter, \ie, the final loss will be either $\textrm{CTC} + \alpha\textrm{CE}$ or $\textrm{CTC} + \alpha \textrm{OT}$. We performed a grid search for $\alpha$ among $[0.1, 0.2, 0.5, 1.0]$ on \mustc En-De and found that $\alpha=0.1$ works well for both CE and OT (the other values yielded similar performance). Therefore, we set $\alpha=0.1$ in all experiments. Note that previous work~\cite{wang2020curriculum,zhang2020adaptive} used $0.7\textrm{CE}+0.3\textrm{CTC}$, but we found that this combination performs worse than our choice (see Table~\ref{tab:ablation_ctc_ce} in Appendix~\ref{sec:appendix_ctc_ce}), which again indicates the importance of CTC. For the Wasserstein loss, we use the efficient GPU implementation by~\citet{feydy2019interpolating} with the default parameters (regularization weight $\lambda = 1$ and $\ell_2$ cost function). Tuning these could further boost the performance, but we use the default for simplicity. For the weight $\gamma$ of OT positional encoding (see~\eqref{eq:positional_ot}), after doing a quick grid search among $[0.1, 0.5, 1.0, 5.0, 10.0]$ on a small training subset of \mustc En-De, we found that $\gamma = 1.0$ works reasonably
well and thus we use this value in all experiments.

\subsection{Ablation analysis}\label{sec:ablation-analysis}
We first conduct ablation studies to support some of our claims and to validate the rationale behind various design choices for our method. The following results are obtained under the \emph{bilingual} setting using two language pairs (En-De and En-Fr) of the \mustc \emph{dev} sets, and \emph{with-MT} setting. More ablation results can be found in Appendix~\ref{sec:ablation_more}.

\paragraph{ASR \vs ST performance}
The results shown in Table~\ref{tab:ablation_bleu_vs_wer} strongly support our claim that CTC obtains better final ST accuracy than CE (and not far off from CTC+CE) despite its inferior ASR performance (recall our claim that CTC can reduce modality gap). Comparing CTC and CTC+OT, we see that OT could only improve WER very slightly ($0.1$-$0.2$ points), whereas the gain in the final ST performance is more important ($0.4$-$0.7$ points). This indicates that OT has a bigger impact on the final ST performance than merely improving WER during ASR pre-training.

\begin{table}[!htb]
    \centering%
\caption{\label{tab:ablation_bleu_vs_wer}BLEU and WER (in parentheses) on \mustc \emph{dev} sets. %
}%
\resizebox{\linewidth}{!}{
        \begin{tabular}{l ccc}
        \toprule
            \textbf{Method} & \textbf{En$\to$De} & \textbf{En$\to$Fr} & \textbf{Training time}*\\
        \midrule
        CE & 23.04 \perate{13.1} & 28.89 \perate{12.6} & 7.10h \\
        CTC & 24.08 \perate{17.7} & 29.91 \perate{17.3} & 6.27h\\
        CTC+CE & 24.28 \textbf{\perate{12.9}} & 30.21 \textbf{\perate{11.9}} & 7.78h\\
        CTC+OT & \textbf{24.74} \perate{17.6} & \textbf{30.31} \perate{17.1} & 7.62h\\
        \bottomrule
             
        \end{tabular}
    }
   {\small
    *100 epochs on En-Fr, batch size 40K tokens, 8 V100 GPUs}
\end{table}

\begin{table}[!htb]
\addtolength{\tabcolsep}{-0.2em}
    \centering%
    \caption{BLEU and WER (in parentheses) on \mustc \emph{dev} sets %
    for different distance functions in Siamese pre-training.\label{tab:distance_metrics}}%
    \resizebox{\linewidth}{!}{
        \begin{tabular}{ll c c}
        \toprule
        \textbf{Metric} & \textbf{Length-match}  & \textbf{En$\to$De} & \textbf{En$\to$Fr} \\

        \toprule

        (CTC alone) & - & 24.08 \perate{17.7} & 29.91 \perate{17.3}\\
        \midrule
        
        \multirow{3}{*}{Euclidean} & {average} & 24.41 \perate{18.1} & 29.86 \perate{17.3}\\
        
        & {attention} & 24.30 \perate{18.6} & 29.81 \perate{19.2} \\
        & {interpolation} & 23.94 \perate{19.3} & 29.77 \perate{17.8} \\
        \midrule
        
        \multirow{2}{*}{KL-diverg.} & {attention} & 24.56 \perate{18.3} & 30.10 \perate{\textbf{17.1}} \\
        
        & {interpolation} & 24.26 \perate{18.1} & 29.96 \perate{17.4}\\
        \midrule
        
         Adversarial & - & 23.73 \perate{20.6} & 29.98 \perate{19.6}\\
        \midrule
        
        Wasserstein & - & \textbf{24.74} \perate{\textbf{17.6}} & \textbf{30.31} \perate{\textbf{17.1}} \\
        
        \bottomrule
        \end{tabular}
    }
\end{table}

\paragraph{OT \vs other distances}
As discussed in Section~\ref{sec:learning-to-align}, our Siamese pre-training can use different distance functions than OT (Appendix~\ref{sec:appendix-distance-functions}), including Euclidean distance, KL divergence, Soft-DTW~\cite{cuturi2017soft}, or adversarial loss~\cite{ganin2016domain,lample2018unsupervised,alinejad2020effectively}. The first two require a length-matching operation to be applied to the sequences, for which we experiment with average pooling~\cite{liu2020bridging,dong2021listen}, cross-attention~\cite{tang2021improving}, and linear interpolation (ours). Soft-DTW is excluded due to its prohibitive memory footprint. For all methods we perform the same hyper-parameter search (Section~\ref{sec:experimental-setup}). Table~\ref{tab:distance_metrics} shows that Euclidean distance and adversarial loss do not always improve over CTC,\footnote{Our results of adversarial loss differs from~\citet{alinejad2020effectively} but in line with~\citet{lample2018unsupervised}. See \href{https://github.com/facebookresearch/UnsupervisedMT/issues/40}{\texttt{github.com/facebookresearch/UnsupervisedMT/issues/40}}.
} while KL divergence and OT consistently improve over that baseline, with OT being the best.

\paragraph{Different variants of Siamese pre-training}
As discussed in Section~\ref{sec:proposed-recipe}, our proposed method can have different variants: weight sharing between the encoders,\footnote{The speech encoder has $6$ layers more than the text one, so only the last $6$ are shared.} and using (or not) positional encoding for OT (Section~\ref{sec:learning-to-align}). From the results in Table~\ref{tab:ablation_ot}, we observe that, without MT pre-training, sharing the encoders tends to give slightly better results, while it is the opposite with MT pre-training (but with a larger margin). %
Using OT positional encoding gives better results in most cases. Therefore, we disable weight sharing and use positional encoding in the remaining experiments.
\begin{table}[!htb]
    \centering\small%
    \caption{\label{tab:ablation_ot}BLEU for different variants of Siamese-PT.}
    \begin{tabular}{l cc  c  c}
    \toprule
     & \textbf{Shared} & \textbf{Positional} & \textbf{En-De} & \textbf{En-Fr} \\
     \midrule
    \multirow{4}{*}{
    	\rotatebox[origin=c]{90}{w/o MT}
    	} 
     & - & - & 23.69 & 29.69 \\
     & \checkmark & - & \textbf{24.12} & 29.73 \\
     & - & \checkmark & 23.91 & \textbf{29.83} \\
     & \checkmark & \checkmark & 23.89 & \textbf{29.86} \\
     \midrule
     \multirow{4}{*}{
     	\rotatebox[origin=c]{90}{with MT}
     	} 
     & - & - & 24.29 & 30.04 \\
     & \checkmark & - & 23.99 & 29.68 \\
     & - & \checkmark & \textbf{24.74} & \textbf{30.31} \\
     & \checkmark & \checkmark & 24.29 & 29.64 \\
    \bottomrule
    \end{tabular}\vspace{-10pt}
\end{table}

\begin{figure*}[!htb]
	\begin{subfigure}[b]{0.325\linewidth}%
	\resizebox{1.0\linewidth}{!}{
			\begin{tikzpicture}

\definecolor{color0}{rgb}{1,0.498039215686275,0.0549019607843137}
\definecolor{color1}{rgb}{0.12156862745098,0.466666666666667,0.705882352941177}
\definecolor{color2}{rgb}{0.172549019607843,0.627450980392157,0.172549019607843}
\definecolor{color3}{rgb}{0.83921568627451,0.152941176470588,0.156862745098039}

\begin{axis}[
legend cell align={left},
legend style={fill opacity=0.8, draw opacity=1, text opacity=1, draw=white!80!black},
tick align=outside,
tick pos=left,
x grid style={white!69.0196078431373!black},
xmin=-0.3, xmax=6.3,
xtick style={color=black},
xtick={0,1,2,3,4,5,6},
xticklabels={ru,pt,de,ca,fr,it,es},
y grid style={white!69.0196078431373!black},
ymin=-0.723, ymax=3.743,
ytick style={color=black},
typeset ticklabels with strut,  %
legend columns=4,
transpose legend
]

\addplot [ultra thick, color0]
table {%
0 0.76
1 0.44
2 0.69
3 0.36
4 0.53
5 -0.01
6 0.25
};
\addlegendentry{CEmt}
\addplot [ultra thick, color1]
table {%
0 1.3
1 1.72
2 1.35
3 0.73
4 0.48
5 0.41
6 0.23
};
\addlegendentry{CTCmt}
\addplot [ultra thick, color2]
table {%
0 2.72
1 1.98
2 0.93
3 0.65
4 0.18
5 0.02
6 -0.05
};
\addlegendentry{CTC+CEmt}
\addplot [ultra thick, color3]
table {%
0 2.79
1 2.74
2 1.82
3 1.67
4 1.56
5 1.14
6 0.99
};
\addlegendentry{CTC+OTmt}
\addplot [ultra thick, color0, dashed]
table {%
0 0
1 0
2 0
3 0
4 0
5 0
6 0
};
\addlegendentry{CE}
\addplot [ultra thick, color1, dashed]
table {%
0 1.59
1 1.53
2 0.81
3 0.37
4 0.09
5 0.03
6 -0.05
};
\addlegendentry{CTC}
\addplot [ultra thick, color2, dashed]
table {%
0 3.27
1 1.45
2 0.35
3 0
4 -0.04
5 -0.52
6 -0.18
};
\addlegendentry{CTC+CE}
\addplot [ultra thick, color3, dashed]
table {%
0 3.54
1 2.05
2 1.36
3 1.04
4 0.78
5 0.88
6 0.69
};
\addlegendentry{CTC+OT}
\end{axis}

\end{tikzpicture}
		}
		\caption{\label{fig:covost-many-to-one}\covostt X*$\to$En.}
	\end{subfigure} %
	\begin{subfigure}[b]{0.325\linewidth}
	\resizebox{1.007\linewidth}{!}{
		\begin{tikzpicture}

\definecolor{color0}{rgb}{1,0.498039215686275,0.0549019607843137}
\definecolor{color1}{rgb}{0.12156862745098,0.466666666666667,0.705882352941177}
\definecolor{color2}{rgb}{0.172549019607843,0.627450980392157,0.172549019607843}
\definecolor{color3}{rgb}{0.83921568627451,0.152941176470588,0.156862745098039}

\begin{axis}[
legend cell align={left},
legend style={fill opacity=0.8, draw opacity=1, text opacity=1, draw=white!80!black},
tick align=outside,
tick pos=left,
x grid style={white!69.0196078431373!black},
xmin=-0.6, xmax=12.6,
xtick style={color=black},
xtick={0,1,2,3,4,5,6,7,8,9,10,11,12},
xticklabels={cy,ca,sl,de,id,ar,sv,tr,lv,et,ta,fa,mn},
y grid style={white!69.0196078431373!black},
ymin=-0.723, ymax=3.743,
ytick style={color=black},
typeset ticklabels with strut,  %
]
\addplot [ultra thick, color0]
table {%
0 0.899999999999999
1 0.899999999999999
2 1
3 0.82
4 0.640000000000001
5 0.9
6 0.720000000000002
7 0.77
8 0.619999999999999
9 0.699999999999999
10 0.26
11 0.58
12 0.369999999999999
};
\addplot [ultra thick, color1]
table {%
0 1.88
1 1.97
2 1.77
3 1.78
4 1.79
5 1.88
6 1.62
7 1.4
8 1.5
9 1.39
10 1.31
11 1.24
12 1.16
};
\addplot [ultra thick, color2]
table {%
0 2.15
1 2.08
2 2.12
3 1.75
4 1.87
5 1.95
6 1.73
7 1.71
8 1.8
9 1.58
10 1.59
11 1.38
12 1.15
};
\addplot [ultra thick, color3]
table {%
0 3
1 2.98
2 2.6
3 2.6
4 2.59
5 2.57
6 2.38
7 2.36
8 2.15
9 2.14
10 1.89
11 1.84
12 1.81
};
\addplot [ultra thick, color0, dashed]
table {%
0 0
1 0
2 0
3 0
4 0
5 0
6 0
7 0
8 0
9 0
10 0
11 0
12 0
};
\addplot [ultra thick, color1, dashed]
table {%
0 1.49
1 1.67
2 1.39
3 1.42
4 1.33
5 1.42
6 1.35
7 1.24
8 0.979999999999999
9 0.99
10 0.93
11 0.9
12 0.890000000000001
};
\addplot [ultra thick, color2, dashed]
table {%
0 1.65
1 1.61
2 1.66
3 1.4
4 1.39
5 1.46
6 1.36
7 1.32
8 1.32
9 1.1
10 0.92
11 1.14
12 1.03
};
\addplot [ultra thick, color3, dashed]
table {%
0 2.58
1 2.29
2 1.98
3 2.14
4 2.11
5 2.15
6 1.99
7 1.84
8 1.38
9 1.62
10 1.59
11 1.53
12 1.5
};
\end{axis}

\end{tikzpicture}
		}
		\caption{\label{fig:covost-one-to-many}\covostt En$\to$X.}
	\end{subfigure}%
	\begin{subfigure}[b]{0.35\linewidth}
	\resizebox{1.0\linewidth}{!}{
		\begin{tikzpicture}

\definecolor{color0}{rgb}{1,0.498039215686275,0.0549019607843137}
\definecolor{color1}{rgb}{0.12156862745098,0.466666666666667,0.705882352941177}
\definecolor{color2}{rgb}{0.172549019607843,0.627450980392157,0.172549019607843}
\definecolor{color3}{rgb}{0.83921568627451,0.152941176470588,0.156862745098039}

\begin{axis}[
legend cell align={left},
legend style={fill opacity=0.8, draw opacity=1, text opacity=1, at={(0.03,0.03)}, anchor=south west, draw=white!80!black},
tick align=outside,
tick pos=left,
x grid style={white!69.0196078431373!black},
xmin=-0.35, xmax=7.35,
xtick style={color=black},
xtick={0,1,2,3,4,5,6,7},
xticklabels={de,es,it,nl,pt,ru,fr,ro},
y grid style={white!69.0196078431373!black},
ymin=-0.691999999999998, ymax=2.272,
ytick style={color=black},
typeset ticklabels with strut,  %
]
\addplot [ultra thick, color0]
table {%
0 0.419999999999998
1 0.449999999999996
2 0.52
3 0.27
4 0.279999999999998
5 -0.0499999999999989
6 0.32
7 -0.579999999999998
};
\addplot [ultra thick, color1]
table {%
0 1.54
1 1.22
2 1.06
3 1
4 1.24
5 0.82
6 1
7 0.900000000000002
};
\addplot [ultra thick, color2]
table {%
0 1.14
1 1.63
2 0.919999999999998
3 1.15
4 1.35
5 1.13
6 1.01
7 0.860000000000003
};
\addplot [ultra thick, color3]
table {%
0 1.66
1 1.51
2 1.6
3 1.48
4 1.28
5 0.870000000000001
6 0.970000000000002
7 0.650000000000002
};
\addplot [ultra thick, color0, dashed]
table {%
0 0
1 0
2 0
3 0
4 0
5 0
6 0
7 0
};
\addplot [ultra thick, color1, dashed]
table {%
0 1.29
1 1.27999999999999
2 1.17
3 1.15
4 1.19
5 1.09
6 0.82
7 0.73
};
\addplot [ultra thick, color2, dashed]
table {%
0 1.45
1 1.41999999999999
2 1.07
3 1.18
4 0.859999999999999
5 0.85
6 1
7 0.540000000000003
};
\addplot [ultra thick, color3, dashed]
table {%
0 1.38
1 1.44
2 1.35
3 1.22
4 1.4
5 0.85
6 1.09
7 0.710000000000001
};
\end{axis}

\end{tikzpicture}
		}
		\caption{\label{fig:mustc}\mustc.}
	\end{subfigure}%
	\caption{\label{fig:results-dev-sets}Results on the \emph{dev} sets, relative to CE. The postfix ``mt'' in the legends indicates that \emph{MT pre-training} was performed (see Figures~\ref{fig:mt-pre-training} and~\ref{fig:mt_ours}). Best viewed in color. Detailed results can be found in Tables~\ref{tab:covost_o2m_dev}--\ref{tab:mustc_dev} in Appendix~\ref{sec:appendix-detailed-results}.%
	}%
\end{figure*}
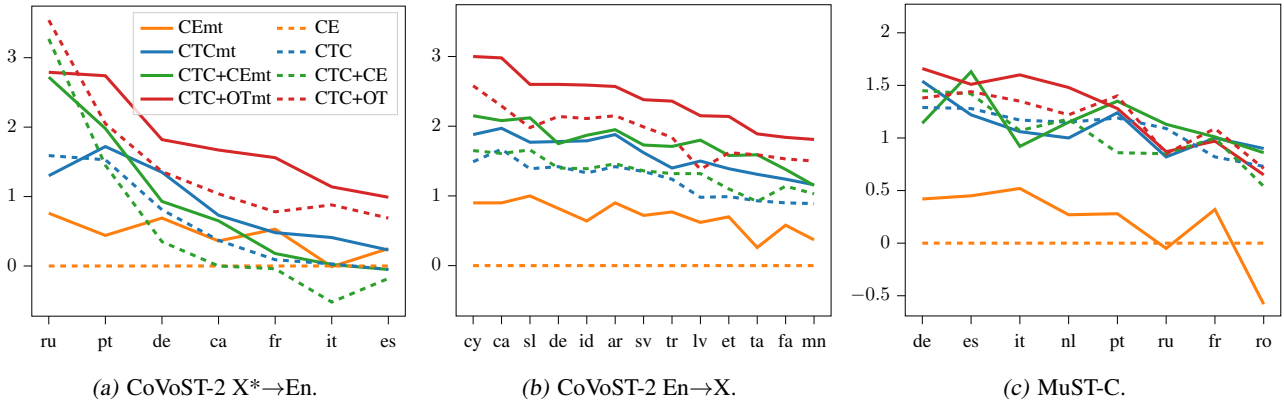

\subsection{Comparative results on the dev sets}\label{sec:results-dev-set}
In this section, we compare the performance of the presented pre-training methods (CE, CTC, CTC+CE, and CTC+OT). A preliminary comparison has appeared in the previous section (Table~\ref{tab:ablation_bleu_vs_wer}), but under a restricted setting (\emph{bilingual} and \emph{with-MT}). Here we provide more extensive results where we also consider the \emph{multilingual} (our focus) and \emph{without-MT} settings. As the number of configurations is still high at this point, we keep using the \emph{dev} sets instead of the \emph{test} sets for performance evaluation. Figure~\ref{fig:results-dev-sets} presents the results in terms of relative BLEU with respect to the baseline CE, and Table~\ref{tab:dev_short} shows absolute BLEU. For the very low-resource languages in \covostt many-to-one (X$\to$En), the results are very poor as no external data is used (\numrange{0.13}{4.56} points, see Table~\ref{tab:covost-many-to-one-dev} in Appendix~\ref{sec:appendix-detailed-results}), which makes the variance too high to draw comparisons.
Therefore, we exclude these languages from the presentation, indicated by an asterisk: X*$\to$En (see Table~\ref{tab:covost-many-to-one-dev} for complete results). We observe that CTC outperforms CE while being competitive with CTC+CE. In general, CTC+OT is the best performing method and is competitive with all other recipes even without MT pre-training.

\begin{table}[htb!]
    \centering%
    \caption{BLEU on the \emph{dev} sets. The prefixes ``(b)'' and ``(m)'' mean bilingual and multilingual (on average). See Tables~\ref{tab:covost_o2m_dev}--\ref{tab:mustc_dev} in Appendix~\ref{sec:appendix-detailed-results} for detailed results.%
    \label{tab:dev_short}}%
    \addtolength{\tabcolsep}{-0.2em}
    \resizebox{\linewidth}{!}{
    \begin{tabular}{l l cc c c c}
        \toprule
        &\multirow{2}{*}{\textbf{Method}} & \multicolumn{3}{c}{\textbf{\mustc}} & \multicolumn{2}{c}{\textbf{\covostt}}\\
        \cmidrule(lr){3-5}\cmidrule(l){6-7}
        &  & (b)\textbf{De} & (b)\textbf{Fr} &
         (m)\textbf{avg} & \textbf{En$\to$X} & \textbf{X*$\to$En}\\
\midrule
         \multirow{4}{*}{\rotatebox[origin=c]{90}{w/o MT}} & CE & 22.39 & 28.48 & 25.42 & 20.50 & 19.64 \\
         & CTC & 23.25 & \textbf{29.87} & 26.51 & 21.71 & 20.27 \\
         & CTC+CE & 23.77 & 29.74 & 26.47 & 21.83 & 20.26  \\
         & CTC+OT & \textbf{23.91} & 29.83 &  \textbf{26.60} & \textbf{22.36} & \textbf{21.12}  \\
         \midrule
         \multirow{4}{*}{\rotatebox[origin=c]{90}{with MT}} & CE & 23.04 & 28.89 & 25.62 & 21.17 & 20.07 \\
         & CTC & 24.08 & 29.91 & 26.52 & 22.10 & 20.53 \\
         & CTC+CE & 24.28 & 30.21 & 26.57 & 22.22 & 20.56 \\
         & CTC+OT & \textbf{24.74} & \textbf{30.31} & \textbf{26.67} & \textbf{22.82} & \textbf{21.46} \\
         \bottomrule
    \end{tabular}
    }\vspace{-5pt}
\end{table}

\subsection{Comparison to state-of-the-art methods}
\label{sec:test_results}
In this section, we compare our \emph{multilingual} models (with MT pre-training) to state-of-the-art methods. Recall that we only use a simple \emph{vanilla} encoder-decoder architecture. Tables~\ref{tab:covost_test} and~\ref{tab:mustc_test_results} show the results on \covostt and \mustc (\emph{test} sets), respectively. For reference, we name our proposed pre-training method \mbox{\textbf{Siamese-PT}} (CTC+OT). We should note that due to the low-resource nature of \covostt, a comparison to methods using external data would be highly unfair. Therefore, we only compare against~\citet{wang2020covost2} whose results were obtained under the same settings as ours. Our proposed method can certainly use external data, which is left for future work.

We observe the effectiveness of CTC pre-training from the \mustc results in Table~\ref{tab:mustc_test_results}. It surpasses CE by $0.4$ points ($29.2$ \vs $28.8$; for the medium model the gap is $1.2$ points, see Table~\ref{tab:mustc_test_results_full}); and adding CE to CTC did not help. Siamese pre-training with OT helps improve BLEU score
to $29.8$. Moreover, we have improved from the previous best results in the no-external-data and multilingual settings~\cite{le2021lightweight} (4\textsuperscript{th} row) by $3.2$ points ($29.8$ \vs $26.6$). Finally, 
our best result even surpasses previous strong and sophisticated multi-task learning systems~\cite{tang2021improving,ye2022cross} that were trained on external data. %
The effectiveness of CTC and OT is again confirmed on \covost, as shown in Table~\ref{tab:covost_test}, where Siamese-PT also achieved the best results.

\begin{table}[htb!]
    \centering
    \caption{BLEU on \covostt \emph{test} set. ``X**$\to$En'' denotes the average excluding very low-resource languages (see Section~\ref{sec:results-dev-set}) %
    and those not reported in~\citet[Table 3]{wang2020covost2}. See 
    Tables~\ref{tab:covost_o2m_test} and~\ref{tab:covost_m2o_test} (Appendix~\ref{sec:appendix-detailed-results}) for details. %
    The reported results are obtained using large model for En$\to$X and medium model for X**$\to$En, which is better than the large one~\cite{wang2020covost2}.\label{tab:covost_test}}%
    \small
    \resizebox{0.85\linewidth}{!}{
    \begin{tabular}{l cc}
        \toprule
        \textbf{Method} & \textbf{En$\to$X} & \textbf{X**$\to$En}\\
         \toprule
         \citet{wang2020covost2} & 19.4 & 24.5\\
         CE & 19.2 & 24.6 \\
         CTC & 19.8 & 24.7 \\
         CTC+CE & 19.7 & 24.5 \\
         Siamese-PT (this work) & \textbf{21.5} & \textbf{25.5} \\
         \bottomrule
    \end{tabular}
    }
    \vspace{-5pt}
\end{table}

\begin{table*}[t]
\addtolength{\tabcolsep}{-0.2em}
\centering
\small
\caption{\label{tab:mustc_test_results}Performance on \mustc test set. Due to space constraints, only the \emph{large} model and only some existing methods are included. Our results for the \emph{medium} model are also competitive. See Table~\ref{tab:mustc_test_results_full} in Appendix~\ref{sec:appendix-detailed-results} for a complete list. The column ``Multi'' denotes multilingual models (instead of individual bilingual ones). %
}%
\resizebox{\linewidth}{!}{
      \begin{tabular}{lrc cc cccc cccc  a}
		\toprule
        \multicolumn{2}{l}{\multirow{2}{*}{\textbf{Method}}} & \multirow{2}{*}{\textbf{Multi}} & \multicolumn{2}{c}{\textbf{External Data}} & \multicolumn{9}{c}{\textbf{BLEU}} \\
		\cmidrule(lr){4-5}\cmidrule(l){6-14}
		& & & \textit{Unlabeled} & \textit{Labeled} & \textbf{de} & \textbf{es} & \textbf{fr} & \textbf{it} & \textbf{nl} & \textbf{pt} & \textbf{ro} & \textbf{ru} & \textbf{avg} \\ 
		
		\midrule
		
		\multicolumn{2}{l}{\textsc{fairseq S2T}~\cite{wang2020fairseq}}& \checkmark& - & - & 
		24.5 & 28.2 & 34.9 & 24.6 & 28.6 & 31.1 & 23.8 & 16.0 & 26.5 \\
		
		\multicolumn{2}{l}{ESPnet-ST~\cite{inaguma2020espnet}} & \checkmark  & -  & - & 
		22.9 & 28.0 & 32.7 & 23.8 & 27.4 & 28.0 & 21.9 & 15.8 & 25.1  \\

		\multicolumn{2}{l}{Dual-decoder~\cite{le2020dual}} & \checkmark  & - & - & 
		23.6 &28.1 & 33.5 & 24.2 & 27.6 & 30.0 & 22.9 & 15.2 & 25.6 \\
		
		\multicolumn{2}{l}{Adapters~\cite{le2021lightweight}} & \checkmark & - & -    & 
		24.7 & 28.7 & 35.0 & 25.0 &  28.8 & 31.1 & 23.8 & 16.4 & 26.6 \\

		\multicolumn{2}{l}{BiKD~\citep{inaguma2021source}} & - & - & - & 
		25.3 & - & 35.3 & - & - & - & - & -  & - \\

		\multicolumn{2}{l}{JointSpeechText~\cite{tang2021improving}} & - & - & \checkmark &
		26.8 & 31.0 & 37.4 &- &-&-&-& - & - \\

		\multicolumn{2}{l}{TaskAware~\cite{indurthi2021task}} & - & - & \checkmark  &
		\textbf{28.9}& - &- &-&-&-& - & - & - \\

		\multicolumn{2}{l}{ConST~\cite{ye2022cross}}  & - & \checkmark &  \checkmark  &
		28.3  & 32.0 & 38.3 & 27.2 & \textbf{31.7} & 33.1 & 25.6 & \textbf{18.9} & 29.4 \\ 
		
		\multicolumn{2}{l}{STPT~\cite{tang2022unified} }& - & \checkmark & \checkmark & 
		- & \textbf{33.1} & \textbf{39.7} &-&-&-&-& - & -\\
		
		\midrule

		CE pre-training & \multirow{4}{*}{\textsc{medium}} & \checkmark & - & - & 24.6 & 28.7 & 34.9 & 24.6 & 28.4 & 30.7 & 23.7 & 15.9 & 26.4 \\
		CTC pre-training&  & \checkmark & - & - & 25.9 & 29.7 & 36.6 & 25.6 & 29.6 & 32.0 & 24.6 & 16.7 & 27.6 \\
		CTC+CE pre-training & & \checkmark & - & - & 25.6 & 29.5 & 36.4 & 25.2 & 29.5 & 31.6 & 24.5 & 16.5 & 27.4 \\
		Siamese-PT (this work)& 
		& \checkmark & - & - & 26.2 & 29.8 & 36.9 & 25.9 & 29.8 & 32.1 & 24.8 & 16.8 & 27.8 \\
		\midrule

		CE pre-training&  \multirow{4}{*}{\textsc{large}}& \checkmark & - & - & 26.9 & 30.8 & 37.7 & 26.7 & 30.8 & 33.3 & 26.2 & 17.9 & 28.8 \\
		
		CTC pre-training& & \checkmark & - & - & 27.6 & 31.4 & 38.2 & 27.2 & 31.1 & 33.6 & 26.4 & 18.4 & 29.2 \\
		
 		CTC+CE pre-training& & \checkmark & - & - & 27.2 & 31.2 & 38.0 & 27.0 & 31.5 & 33.7 & 26.2 & 18.3 & 29.1 \\

        Siamese-PT (this work)& & \checkmark & - & - & 27.9 & 31.8 & 39.2 & \textbf{27.7}& \textbf{31.7} & \textbf{34.2} & \textbf{27.0} & 18.5 & \textbf{29.8} \\
		\bottomrule
	\end{tabular}
}\vspace{-5pt}
\end{table*}

\subsection{Application to multi-task learning}\label{sec:multi-task}

As a proof of concept of the wide applicability of our method, we apply it on top of the multi-task learning system by~\citet{tang2021improving}, where %
they pre-trained the ASR component with %
CE loss. We replicate their results
on \mustc En-De and perform further experiments using different ASR pre-training methods discussed in the paper. %
The results are shown in Table~\ref{tab:joint_speech_text_results}. We observe again the superiority of CTC over CE. Simply replacing CE with CTC yields an improvement of $0.26$ points, whereas using them together slightly decreases the performance. Our Siamese-PT reached the highest accuracy with an improvement of $0.46$, %
showing that our pre-training method is not only effective for vanilla ST but also complementary to multi-task learning. \vspace{-5pt} 

\begin{table}[!htb]
\caption{\label{tab:joint_speech_text_results}Results of the multi-task learning system of~\citet{tang2021improving}, using different ASR pre-training methods. %
}%
    \centering\small
    \resizebox{0.78\linewidth}{!}{
        \begin{tabular}{lc}
    \toprule
         \textbf{ASR pre-training method} & \textbf{En-De} \\    
    \midrule
    CE (reported in~\citet{tang2021improving}) & 26.74 \\
    CE (reproduced) & 26.78 \\
    CTC & 27.04 \\
    CTC+CE & 26.69 \\
    Siamese-PT (CTC+OT, proposed) & \textbf{27.20} \\
    \bottomrule
    \end{tabular}%
    }
\end{table}

\vspace{-10pt}
\section{Conclusion}\label{sec:conclusion}
In this paper, we have discussed the gap between speech and text modalities as a challenge in speech-to-text translation, and proposed to overcome it at the pre-training stage. We have shown that ASR pre-training using the CTC loss can reduce this modality gap, but only partially. To further mitigate this issue, we have proposed a novel pre-training method in which we train a Siamese-like model composed of a speech encoder and a text encoder, such that they produce representations that are close in the Wasserstein space. Extensive experiments on two popular ST datasets have demonstrated the effectiveness of our method. In particular, our results surpassed previous work %
in the non-external-data setting, while being on par with recent strong multi-task learning systems %
trained with external data. %
Our pre-training method can also be applied to the multi-task learning system, yielding further improvements for these methods. %

\paragraph{Limitations and future work} %
Even though the proposed pre-training recipe (Section~\ref{sec:proposed-recipe}) can use additional ASR and MT data, we have only considered the setting without external data. While this setting is important in many real-world scenarios, by restricting to it we have missed the opportunity to show the effectiveness of the proposed method on very low-resource languages, as discussed in Section~\ref{sec:results-dev-set}. On the other hand, the proposed method for learning to align speech and text features (Section~\ref{sec:learning-to-align}) is quite generic and has potential applications beyond pre-training for ST. For example, the proposed OT speech-text alignment could be, in principle, used as an additional loss in multi-task learning. We aim to address all these open questions in future work.\vspace{-5pt}

\paragraph{Societal impact} The technology developed in this work can be used in surveillance systems, which might raise privacy concerns. Therefore, considerations should be taken when releasing or deploying these models to the public.

\section*{Acknowledgements}
This work was supported by a Meta AI SRA grant and partially by the Multidisciplinary Institute in Artificial Intelligence (MIAI@Grenoble Alpes, ANR-19-P3IA-0003). It was granted access to the HPC/AI resources of IDRIS under the allocations 2022-AD011012538 and 2022-AD011013744 made by GENCI. We thank Yun Tang and Anne Wu for their help in reproducing the results reported in \citet{tang2021improving} and \citet{wang2020covost2}, respectively. We also thank Aidan Mannion and the anonymous reviewers for their feedback that helped greatly improve the manuscript. 

\bibliography{refs}
\bibliographystyle{icml2023}

\appendix

\section{Distance Functions for Speech-Text Alignment}\label{sec:appendix-distance-functions}
We provide a more detailed description of alternative loss functions (other than OT) for learning to align speech and text features, as discussed in Section~\ref{sec:learning-to-align}. Recall that our notations for the output speech and text features are $\U = (\u_1,\dots,\u_m)\in\RR^{d\times m}$ and $\V = (\v_1,\dots,\v_n)\in\RR^{d\times n}$, respectively, where $\u_i\in\RR^d$ and $\v_j\in\RR^d$ for all $i,j$. %
As some distance functions require the sequence lengths to be the same (which is not the case, recall that typically $m\gg n$), we first present methods for making this happen. 

\subsection{Length-matching operations}

In general, the sequences $\U$ and $\V$ are processed to obtain new sequences $\tilde{\U}\in\RR^{d\times p}$ and $\tilde{\V}\in\RR^{d\times p}$ that have the same length $p$ (which needs not equal $m$ or $n$):
\begin{equation}
    \tilde{\U},\tilde{\V} = \textsc{LengthMatching}(\U,\V).
\end{equation}
We have experimented with the following three length-matching operations.

\paragraph{Average pooling} This consists in simply taking the global average of each sequence~\cite{liu2020bridging,dong2021listen}:
\begin{equation}
    \tilde{\U} = \frac{1}{m} \sum_{i=1}^m \u_i,\qquad
    \tilde{\V} = \frac{1}{n} \sum_{j=1}^n \v_j.
\end{equation}
    Note that before using this averaging operation, one may also apply some kind of \emph{shrinking} mechanisms to reduce the length of the speech representations such as using CTC labels as guidance to remove contiguous blanks or repeated predictions~\cite{liu2020bridging}, or using a shrinking layer, for example a 2D convolutional layer followed by a normalization one~\cite{dong2021listen}. These shrinking operations do not necessarily make the length of the speech output equal to that of the text one. For simplicity, we did not experiment with any type of shrinking mechanisms.

\paragraph{Interpolation} This consists in resizing the longer sequence to the same length as the shorter one:%
    \begin{equation}
    \tilde{\U} = \textsc{Interpolate}(\U, n),\qquad
    \tilde{\V} = \V,\qquad
\end{equation}
The function $\textsc{Interpolate}(\cdot,\cdot)$ takes as inputs the sequence $\U$ (of length $m$) together with the desired dimension $n$ and produces the interpolated sequence $\tilde{\U}$ of length $n$. The most popular method of interpolation is perhaps \emph{linear} interpolation, which we use in our experiments, but more advanced ones such as \emph{cubic} interpolation can also be used. In our implementation, we used PyTorch's  \href{https://pytorch.org/docs/stable/generated/torch.nn.functional.interpolate.html}{\texttt{nn.functional.interpolate}} function (with argument \texttt{mode=`linear'}). Since interpolation is a rather standard operation, we omit further details.

\paragraph{Attention-based transformation} This was proposed by~\citet{tang2021improving} under the name Cross-Attentive Regularization (CAR), which consists in using cosine attention~\cite{graves2014neural} to obtain:
\begin{align}
    \tilde{\U} &= \U \softmax({\bar{\U}}^\top \bar{\V})\in\RR^{d\times n}, \\
    \tilde{\V} &= \V \softmax({\bar{\V}}^\top \bar{\V})\in\RR^{d\times n},
\end{align}
where $\bar{\U} = (\bar{\u}_1,\dots,\bar{\u}_m)$ and $\bar{\V} = (\bar{\v}_1,\dots,\bar{\v}_n)$ are sequences of normalized features:
\begin{align}
    \bar{\u}_{i} &= \frac{\u_{i}}{\norm{\u_{i}}_2}, \quad i=1, \dots, m,\\
    \bar{\v}_{j} &= \frac{\v_{j}}{\norm{\v_{j}}_2}, \quad j=1, \dots, n.
\end{align}

\subsection{Distance functions}
We present four different distance functions: Euclidean distance, Kullback-Leibler (KL) divergence, adversarial loss, and dynamic time warping (DTW). For the first two, we apply one of the length-matching operations presented in the previous section.

\paragraph{Euclidean distance} This is computed as
\begin{equation}
    d_{\mathrm{euc}}(\tilde{\U}, \tilde{\V}) = \|\tilde{\U} - \tilde{\V}\|_2 = \sqrt{\sum_{i=1}^p \norm{\tilde{\u}_{i}-\tilde{\v}_{i}}_2^2},
\end{equation}
where $p$ denotes the length of the sequences resulting from the length-matching mechanism. In particular, $p=1$ for \emph{average pooling} and $p=n$ for \emph{interpolation} or \emph{attention}.

\paragraph{Kullback-Leibler divergence} Let $\a = (a_1,\dots,a_d)$ and $\b = (b_1,\dots,b_d)$ (where $a_i, b_i \in [0, 1]$ and $\1^\top\a = \1^\top\b = 1$) be the masses representing two discrete probability distributions. The KL divergence between these distributions is given by
\begin{equation}
    \mathrm{KL}(\a \parallel \b) = \sum_{k=1}^d a_k\log \frac{a_k}{b_k}.
\end{equation}
For two input sequences $\tilde{\U}\in\RR^{d\times p}$ and $\tilde{\V}\in\RR^{d\times p}$ (already length-matched), we first normalize the features to obtain probability vectors:
\begin{align}
    \bar{\u}_i &= \softmax(\tilde{\u}_i),\quad i=1, \dots, p,\\ \bar{\v}_i &= \softmax(\tilde{\v}_i),\quad i=1, \dots, p.
\end{align}
Then our KL-based distance between the input sequences is given by:
\begin{equation}
    d_{\mathrm{KL}}(\tilde{\U},\tilde{\V}) = \sum_{i=1}^p \mathrm{KL}(\bar{\u}_i \parallel \bar{\v}_i).
\end{equation}

\paragraph{Adversarial loss} The idea is to add a \emph{discriminator} module that learns to distinguish speech features from text features~\cite{ganin2016domain,lample2018unsupervised,alinejad2020effectively} (highly inspired by adversarial generative networks~\cite{goodfellow2014generative}). In our case, the speech and text encoders act as a \emph{generator} that tries to deceive the discriminator. Intuitively, if the discriminator fails to distinguish between the speech and text features, then we can say that the encoders are successful at producing speech and text features that are close to each other. To achieve this, we add two loss functions (in addition to existing ones such as CTC):
\begin{enumerate}
    \item A \emph{discriminator loss} that helps train the discriminator to distinguish between the speech and text features.
    \item A \emph{generator loss} that helps train the generator (\ie, the two encoders) to fool the discriminator.
\end{enumerate}
Then, training alternates between updating the discriminator's parameters (while freezing the generator) and updating the generator's parameters (while freezing the discriminator). More formally, given an input feature vector $\u\in\RR^d$, the discriminator outputs a value $p\in[0,1]$ representing the probability of $\u$ being produced by a \emph{speech} encoder (or equivalently, $1-p$ denotes the probability of $\u$ being produced by the \emph{text} encoder):
\begin{equation}
    p = \textsc{Discriminator}(\u).
\end{equation}
For a pair of sequences $\U = (\u_1,\dots,\u_m)\in\RR^{d\times m}$ and $\V = (\v_1,\dots,\v_n)\in\RR^{d\times n}$ produced by the speech and text encoders, respectively, we compute the discriminator loss as
\begin{equation}
    L_{\mathrm{disc}}(\U,\V) = \sum_{i=1}^m \bce(p_i, 1) \\
    + \sum_{j=1}^n \bce(q_j, 0),
\end{equation}
where
\begin{align}
    p_i &= \textsc{Discriminator}(\u_i), \\
    q_j &= \textsc{Discriminator}(\v_j),
\end{align}
and $\bce(\cdot,\cdot)$ is the binary cross-entropy loss:
\begin{equation}
    \bce(p, y) = -y\log p - (1-y)\log (1-p),
\end{equation}
where $y\in\set{0,1}$ is the ground-truth label.

The generator loss is similar, except that we switch the labels:
\begin{equation}
    L_{\mathrm{gen}}(\U,\V) = \sum_{i=1}^m \bce(p_i, 0) \\
    + \sum_{j=1}^n \bce(q_j, 1).
\end{equation}
We use the same set of hyper-parameters as~\citet{lample2018unsupervised} (which was also used by \citet{alinejad2020effectively}). In particular, our discriminator is a $3$-layer feedforward network with hidden dimension $1024$ and leaky ReLU activations (with negative slope 0.2), each followed by a Dropout layer of probability $0.1$.

\paragraph{Differentiable DTW} Dynamic time warping~\cite{sakoe1978dynamic} is a popular approach to measuring the similarity between two time series, making it a natural choice to the problem of aligning speech and text, both of which are temporal sequences. A differentiable variant of this function has been proposed~\cite{cuturi2017soft} to allow training with stochastic gradient descent. Even though this is also available in our implementation, we did not include it in our experiments due to its prohibitively high memory usage. Therefore, we refer the reader to~\citet{cuturi2017soft} for further details.

\section{Further Implementation Details}\label{sec:implentation-details}

\paragraph{Data processing}
The speech input features are 80-dimensional log Mel filter-bank. Utterances having more than 3000 frames are removed for GPU efficiency. For data augmentation, following \cite{wang2020fairseq,wang2020covost2} we used SpecAugment~\cite{park2019spec} with LibriSpeech basic (LB) policy (no time warping) in the bilingual \mustc and multilingual settings of \covostt. For multilingual \mustc, we applied LibriSpeech double (LD) policy. The inputs to the text encoder in our ASR and MT pre-training phases are in their phoneme pronunciation form, except for the many-to-one case in which we used the 10k unigram vocabulary~\cite{kudo2018sentencepiece} learned on the training text of all source languages. Following~\cite{tang2021improving}, the grapheme-to-phoneme conversion is done using the {g2p\_en} package~\cite{lee2018learning}. The resulting English phoneme vocabulary has the size of 134 tokens. On the target side for MT and ST models, we used 10K unigram vocabulary learned on all relevant target training text. More specifically, the vocabulary is learned on German or French training text in the bilingual case, and on the concatenation of all target training texts in the multilingual ones.

\paragraph{Training hyper-parameters}
We used the Adam optimizer~\cite{kingma2015adam} with learning rate linearly increased for the first $N$ warmup steps to a value $\eta_{\max}$, then decreased proportionally to the inverse square root of the step counter. $\eta_{\max}$ is set to \num{2e-3} in medium ASR/ST experiments and to \num{5e-4} in experiments using large architecture. For MT experiments, $\eta_{\max}=\num{5e-3}$. $N$ is set to $10000$ in ASR/ST experiments and to $4000$ in MT experiments. %
Label smoothing is set to 0.1~\cite{szegedy2016rethinking} for models using cross-entropy loss. All ST models on \mustc are trained up to 50 epochs and those on \covostt are trained up to 50 and 100 epochs for En$\to$X and X$\to$En, respectively. The last 10 checkpoints are averaged for decoding using a beam size of 5.

\paragraph{Metrics} We report the case-sensitive detokenized BLEU scores~\cite{papineni2002bleu,post2018call}, except for English-Japanese and English-Chinese where we report character-level BLEU~\cite{wang2020covost2}. For ASR performence, we report Word Error Rate (WER). 

\paragraph{Hardware and training time} Medium ASR/ST models were trained on 8 NVIDIA V100 GPUs while large ASR/ST ones were trained on 32 A100 GPUs. All MT models were trained on 8 V100 GPUs. %
A comparison of the training time is given in Table~\ref{tab:running_time}.

\begin{table}[!ht]
    \centering
    \caption{\label{tab:running_time}Training time for 100 epochs on \mustc En-Fr (batch size 40K tokens).}
    \begin{tabular}{lcc}
    \toprule
        \textbf{Method} & \textbf{Hours} & \textbf{Slowdown \vs CTC} \\
    \midrule
        CE & 7.10 & 1.13$\times$ \\
        CTC & 6.27 & 1.00$\times$ \\
        CTC+CE & 7.78 & 1.24$\times$ \\
        CTC+OT & 7.62 & 1.22$\times$ \\
    \bottomrule
    \end{tabular}
\end{table}

\section{Further Ablation Studies}\label{sec:ablation_more}

\subsection{CTC encoder \vs CE encoder}\label{sec:appendix-asr_performance_speech_encoders}
In Section~\ref{sec:ctc_reduces_gap} we claim that the speech encoder trained by CTC is stronger than the one trained by CE. The experimental results in Section~\ref{sec:experiments} already support this claim. Indeed, the ST results labeled with ``CTC'' are consistently better than those with ``CE''. Given that the two models are identical except for the encoder initialization, any difference in the ST performance between them can technically be attributed to the quality of the initialized encoders. This allows us to conclude that the CTC encoder is superior to the CE encoder (in terms of producing better embeddings for the subsequent ST task).

In this section, we present additional results on the ASR task to further demonstrate the superiority of the CTC encoder. 

\begin{itemize}
    \item Consider two pre-trained speech encoders, one pre-trained with CE and the other with CTC, as shown in Figures~\ref{fig:asr-pre-training} and~\ref{fig:ctc-pre-training} (without the CE branch), respectively. Previously we have referred to these as ``CE encoder'' and ``CTC encoder'', respectively.
    \item We freeze these encoders, plug them into an encoder-decoder ASR model, and train it with the CE loss.
    \item We do the same as the previous item, but this time with an encoder-only model, trained with the CTC loss. Here we add a linear layer between the encoder and the CTC loss to project the encoder's output hidden dimension to the vocabulary size.
\end{itemize}

The results are shown in Table~\ref{tab:frozen_speech_encoder}. 

\begin{table}[!htb]
    \centering
    \caption{WER on MuST-C En-De \emph{dev} set for comparing CTC encoder and CE encoder. ``Enc'' and ``Dec'' mean ``encoder'' and ``decoder'', respectively.\label{tab:frozen_speech_encoder}}
    \begin{tabular}{lcc}
    \toprule
       \textbf{Frozen Enc}  & \textbf{Enc-Dec-CE} & \textbf{Enc-Linear-CTC}\\
    \midrule
      CE   & 12.7 & 29.9 \\
      CTC & 12.4 & 15.1 \\
    \bottomrule     
    \end{tabular}
\end{table}

We observe that, for the encoder-decoder model, the pre-trained CTC encoder outperforms the pre-trained CE encoder (WER $12.4$ \vs $12.7$) despite the fact that the CE encoder has been previously pre-trained in a similar fashion (using an encoder-decoder model). This provides strong empirical evidence to support our hypothesis that the CTC encoder is superior.

For the encoder-only model, the results are similar, but with a much larger gap (WER $15.1$ for CTC \vs $29.9$ for CE). %
We note that this model has a smaller number of trainable parameters compared to the previous one ($5$M parameters of the linear layer compared to $35$M parameters of the decoder).

\begin{table*}[!htb]
    \centering
    \caption{BLEU on \mustc dev sets for different loss weighting values in CTC+CE.
    \label{tab:ablation_ctc_ce}}
    \resizebox{\linewidth}{!}{
        \begin{tabular}{ll  cc  cccc cccc a}
            \toprule
            & \multirow{2}{*}{\textbf{Method}} & \multicolumn{2}{c}{\textbf{Bilingual}} & \multicolumn{9}{c}{\textbf{Multilingual (one-to-many)}}\\
            \cmidrule(lr){3-4}\cmidrule(l){5-13}
              & & \textbf{de} & \textbf{fr} &
             \textbf{de} & \textbf{es} & \textbf{fr} & \textbf{it} & \textbf{nl} & \textbf{pt} & \textbf{ro} & \textbf{ru} & \textbf{avg} \\
             \midrule
             
             \multirow{2}{*}{w/o MT} & 0.7CE + 0.3CTC & \textbf{24.10} & 29.59 & 24.13 & 32.71 & 29.77 & 24.54 & 26.13 & \textbf{30.24} & 23.27 & 15.86 & 25.83 \\
             & CTC + 0.1CE & {24.09} & \textbf{29.74} & \textbf{25.32} & \textbf{33.80} & \textbf{30.54} & \textbf{25.64} & \textbf{26.97} & 30.07 & \textbf{23.44} & \textbf{15.95} & \textbf{26.47} \\

             \midrule

             \multirow{2}{*}{with MT} & 0.7CE + 0.3CTC & 24.04 & 29.77 & \textbf{25.34} & 33.57 & 30.42 & \textbf{25.61} & \textbf{26.98} & 30.31 & 23.14 & 15.90 & 26.41 \\
             & CTC + 0.1CE & \textbf{24.28} & \textbf{30.21} & 25.01 & \textbf{34.01} & \textbf{30.55} & 25.49 & \textbf{26.94} & \textbf{30.56} & \textbf{23.76} & \textbf{16.23} & \textbf{26.57} \\
             \bottomrule
        \end{tabular}
    }
\end{table*}
\begin{table*}[!t]
\addtolength{\tabcolsep}{-0.2em}
\centering
\caption{BLEU on \covostt dev sets in En$\to$X settings for different choices of tokenization: character (``char'') \vs Sentencepiece 10K unigram (``spm10k'').
\label{tab:ablation_char_spm10k}}
\resizebox{\linewidth}{!}{
\begin{tabular}{l  l ccc ccc ccc ccc ccc  a}
    \toprule
     \textbf{Method} & \textbf{Vocab} & \textbf{ar} & \textbf{ca} & \textbf{cy} & \textbf{de} & \textbf{et} & \textbf{fa} & \textbf{id} & \textbf{ja} & \textbf{lv} & \textbf{mn} & \textbf{sl} & \textbf{sv} & \textbf{ta} & \textbf{tr} & \textbf{zh} & \textbf{avg}\\
     \midrule
     
     \multirow{2}{*}{CE w/o MT} & char & 14.95 & 27.01 & 26.94 & 20.83 & \textbf{14.06} & \textbf{15.77} & 25.07 & 36.04 & 13.92 & 12.31 & \textbf{16.70} & 22.63 & 14.17 & \textbf{13.90} & 30.00 & 20.29 \\
    
      & spm10k & \textbf{15.04} & \textbf{27.23} & \textbf{27.19} & \textbf{20.98} & \textbf{14.06} & 15.76 & \textbf{25.50} & \textbf{36.54} & \textbf{14.06} & \textbf{12.57} & \textbf{16.70} & \textbf{22.95} & \textbf{14.46} & 13.84 & \textbf{30.58} & \textbf{20.50} \\
      
     \midrule
     \multirow{2}{*}{CE w/ MT} & char & 15.86 & 27.99 & 27.92 & 21.71 & 14.53 & 16.25 & 26.05 & 36.89 & \textbf{14.89} & \textbf{12.96} & \textbf{17.77} & 23.51 & \textbf{14.86} & \textbf{14.61} & 30.49 & 21.09 \\
     & spm10k & \textbf{15.94} & \textbf{28.13} & \textbf{28.09} & \textbf{21.80} & \textbf{14.76} & \textbf{16.34} & \textbf{26.14} & \textbf{37.00} & 14.68 & 12.94 & 17.70 & \textbf{23.67} & 14.72 & \textbf{14.61} & \textbf{31.03} & \textbf{21.17}\\
    \bottomrule
\end{tabular}
}

\end{table*}

\subsection{Different loss weighting values in CTC+CE}
\label{sec:appendix_ctc_ce}
The hybrid architecture using CE and CTC has been shown to obtain strong ASR performance~\cite{watanabe2017hybrid,karita2019comparative}. Hence, there are several works using this pre-trained ASR model to initialize subsequent ST models with the combination of 0.7CE + 0.3CTC as proposed in \cite{watanabe2017hybrid,karita2019comparative}. We additionally compare this combination with our choice (CTC + 0.1CE) in order to find the better performing recipe for the CTC+CE pre-training method. Results in Table~\ref{tab:ablation_ctc_ce} show that our combination of CTC+0.1CE works better for the ST initialization in both \emph{with} (improvement of $+0.16$) and \emph{without} (improvement of $+0.64$) MT pre-training settings. The results again highlight the importance of CTC in the ASR pre-training stage.

\subsection{Character \vs SentencePiece tokenization}
\label{sec:ablation-tokenization}
For \covost one-to-many settings, we compare different choices of vocabulary: character \vs 10K unigram vocabulary. The results (see Table~\ref{tab:ablation_char_spm10k}) show that using the 10K unigram slightly outperforms using character dictionary.

\section{Detailed Results and Discussion}\label{sec:appendix-detailed-results}

\begin{table*}[!t]
\addtolength{\tabcolsep}{-0.2em}
\centering
\caption{Detailed BLEU on \covostt dev sets in En$\to$X settings for all directions.
\label{tab:covost_o2m_dev}}
\resizebox{\linewidth}{!}{
\begin{tabular}{ll ccc ccc ccc ccc ccc  a}
    \toprule
     & \textbf{Method} & \textbf{ar} & \textbf{ca} & \textbf{cy} & \textbf{de} & \textbf{et} & \textbf{fa} & \textbf{id} & \textbf{ja} & \textbf{lv} & \textbf{mn} & \textbf{sl} & \textbf{sv} & \textbf{ta} & \textbf{tr} & \textbf{zh} & \textbf{avg}\\
     \midrule

     \multirow{4}{*}{\rotatebox[origin=c]{90}{w/o MT}} & CE & 15.04 & 27.23 & 27.19 & 20.98 & 14.06 & 15.76 & 25.50 & 36.54 & 14.06 & 12.57 & 16.70 & 22.95 & 14.46 & 13.84 & 30.58 & 20.50 \\
     & CTC & 16.46 & 28.90 & 28.68 & 22.40 & 15.05 & 16.66 & 26.83 & 37.74 & 15.04 & 13.46 & 18.09 & 24.30 & 15.39 & 15.08 & 31.57 & 21.71 \\
     & CTC+CE & 16.50 & 28.84 & 28.84 & 22.38 & 15.16 & 16.90 & 26.89 & 38.11 & 15.38 & 13.60 & 18.36 & 24.31 & 15.38 & 15.16 & 31.59 & 21.83 \\
     & CTC+OT & \textbf{17.19} & \textbf{29.52} & \textbf{29.77} & \textbf{23.12} & \textbf{15.68} & \textbf{17.29} & \textbf{27.61} & \textbf{38.21} & \textbf{15.44} & \textbf{14.07} & \textbf{18.68} & \textbf{24.94} & \textbf{16.05} & \textbf{15.68} & \textbf{32.11} & \textbf{22.36} \\
     
     \midrule
     \multirow{4}{*}{\rotatebox[origin=c]{90}{with MT}} & CE & 15.94 & 28.13 & 28.09 & 21.80 & 14.76 & 16.34 & 26.14 & 37.00 & 14.68 & 12.94 & 17.70 & 23.67 & 14.72 & 14.61 & 31.03 & 21.17 \\
     & CTC & 16.92 & 29.20 & 29.07 & 22.76 & 15.45 & 17.00 & 27.29 & 38.03 & 15.56 & 13.73 & 18.47 & 24.57 & 15.77 & 15.24 & 32.02 & 22.10 \\
     & CTC+CE & 16.99 & 29.31 & 29.34 & 22.73 & 15.64 & 17.14 & 27.37 & 38.12 & 15.86 & 13.72 & 18.82 & 24.68 & 16.05 & 15.55 & 32.03 & 22.22 \\
     & CTC+OT & \textbf{17.61} & \textbf{30.21} & \textbf{30.19} & \textbf{23.58} & \textbf{16.20} & \textbf{17.60} & \textbf{28.09} & \textbf{38.63} & \textbf{16.21} & \textbf{14.38} & \textbf{19.30} & \textbf{25.33} & \textbf{16.35} & \textbf{16.20} & \textbf{32.36} & \textbf{22.82} \\
     
    \bottomrule
\end{tabular}
}

\end{table*}

\begin{table*}[!tb]
\addtolength{\tabcolsep}{-0.35em}%
	\centering
    \caption{Detailed BLEU on \covostt dev sets in X$\to$En settings for all directions.
	\label{tab:covost-many-to-one-dev}}
	\resizebox{\linewidth}{!}{
		\begin{tabular}{cl ccc ccc ccc ccc ccc ccc ccc a}
			\toprule
			& \textbf{Method} & \textbf{ar} & \textbf{ca} & \textbf{cy} & \textbf{de} & \textbf{es} & \textbf{et} & \textbf{fa} & \textbf{fr} & \textbf{id} & \textbf{it} & \textbf{ja} & \textbf{lv} & \textbf{mn} & \textbf{nl} & \textbf{pt} & \textbf{ru} & \textbf{sl} & \textbf{sv} & \textbf{ta} & \textbf{tr} & \textbf{zh} & \textbf{avg} \\
			\midrule
			& Data (h) & 2 & 136 & 2 & 184 & 113 & 3 & 49 & 264 & 1 & 44 & 1 & 2 & 3 & 7 & 10 & 18 & 2 & 2 & 2 & 4 & 10 & \\
			\midrule
			\multirow{4}{*}{\rotatebox[origin=c]{90}{w/o MT}} & CE & 1.32 & 22.06 & 3.01 & 19.32 & 25.53 & 1.06 & 3.94 & 27.31 & \textbf{0.68} & 20.18 & 0.35 & 0.78 & \textbf{0.24} & 3.95 & 8.65 & 14.44 & 0.75 & 1.02 & \textbf{0.25} & 3.04 & 7.11 & 7.86 \\
			
			& CTC & 2.97 & 22.43 & 3.26 & 20.13 & 25.48 & 0.94 & 4.05 & 27.40 & 0.30 & 20.21 & 0.39 & 1.50 & \textbf{0.19} & 3.74 & 10.18 & 16.03 & 0.34 & 1.51 & \textbf{0.23} & 3.46 & 7.06 & 8.18\\
			
			& CTC+CE & 2.80 & 22.06 & 2.99 & 19.67 & 25.35 & 0.85 & 3.94 & 27.27 & 0.38 & 19.66 & 0.46 & \textbf{1.65} & \textbf{0.27} & 3.81 & 10.10 & 17.71 & \textbf{0.96} & \textbf{1.87} & \textbf{0.15} & 3.27 & 6.92 &  8.20 \\
			
			& CTC+OT & \textbf{3.50} & \textbf{23.10} & \textbf{4.15} & \textbf{20.68} & \textbf{26.22} & \textbf{1.33} & \textbf{4.56} & \textbf{28.09} & 0.27 & \textbf{21.06} & \textbf{0.89} & 1.31 & \textbf{0.26} & \textbf{4.33} & \textbf{10.70} & \textbf{17.98} & 0.65 & 1.69 & \textbf{0.20} & \textbf{3.63} & 7.18 & \textbf{8.66}\\
			
			\midrule

			\multirow{4}{*}{\rotatebox[origin=c]{90}{with MT}} & CE & 1.72 & 22.42 & 2.85 & 20.01 & 25.78 & 0.75 & 3.66 & 27.84 & 0.22 & 20.17 & 0.22 & 0.86 & 0.14 & 3.70 & 9.09 & 15.20 & 0.32 & 1.63 & 0.19 & 3.00 & 6.90 & 7.94\\
			
			& CTC & 2.20 & 22.79 & 2.90 & 20.67 & 25.76 & \textbf{0.96} & 3.88 & 27.79 & \textbf{0.44} & 20.59 & 0.70 & 1.16 & \textbf{0.19} & \textbf{4.22} & 10.37 & 15.74 & \textbf{0.97} & \textbf{1.68} & \textbf{0.24} & 3.05 & 6.92 & 8.25\\
			
			& CTC+CE & 1.81 & 22.71 & 2.87 & 20.25 & 25.48 & 0.79 & 3.79 & 27.49 & 0.39 & 20.20 &\textbf{0.81} & \textbf{1.66} & \textbf{0.19} & 3.59 & 10.63 & \textbf{17.16} & \textbf{1.12} & 1.46 & \textbf{0.22} & 3.28 & 6.43 & 8.21\\
			
			& CTC+OT & \textbf{3.26} & \textbf{23.73} & \textbf{3.26} & \textbf{21.14} & \textbf{26.52} & \textbf{0.93} & \textbf{4.45} & \textbf{28.87} & \textbf{0.52} & \textbf{21.32} & 0.31 & 1.25 & \textbf{0.25} & \textbf{4.20} & \textbf{11.39} & \textbf{17.23} & \textbf{0.97} & \textbf{1.77} & 0.13 & \textbf{3.53} & 7.19 & \textbf{8.68} \\
			\bottomrule
		\end{tabular}
	}
\end{table*}

\begin{table*}[t]
    \centering
    \caption{Detailed BLEU on \mustc dev sets.
    \label{tab:mustc_dev}}
    \resizebox{0.85\linewidth}{!}{
        \begin{tabular}{l l  cc  cccc cccc  a}
            \toprule
            &\multirow{2}{*}{\textbf{Method}} & \multicolumn{2}{c}{\textbf{Bilingual}} & \multicolumn{9}{c}{\textbf{Multilingual (one-to-many)}}\\
            \cmidrule(lr){3-4}\cmidrule(l){5-13}
            &  & \textbf{de} & \textbf{fr} &
             \textbf{de} & \textbf{es} & \textbf{fr} & \textbf{it} & \textbf{nl} & \textbf{pt} & \textbf{ro} & \textbf{ru} & \textbf{avg} \\
             \toprule
             \multirow{4}{*}{\rotatebox[origin=c]{90}{w/o MT}} & CE & 22.39 & 28.48 & 23.87 & 32.38 & 29.54 & 24.57 & 25.79 & 29.21 & 22.90 & 15.10 & 25.42 \\
             & CTC & 23.25 & \textbf{29.87} & 25.16 & 33.66 & 30.36 & 25.74 & 26.94 & 30.40 & 23.63 & 16.19 & 26.51\\
             & CTC+CE & 23.77 & 29.74 & \textbf{25.32} & {33.80} & 30.54 & 25.64 & 26.97 & 30.07 & 23.44 & \textbf{15.95} & 26.47 \\
             & CTC+OT & \textbf{23.91} & 29.83 & 25.25 & \textbf{33.82} & \textbf{30.63} & \textbf{25.92} & \textbf{27.01} & \textbf{30.61} & \textbf{23.61} & \textbf{15.95} & \textbf{26.60} \\
             \midrule
             \multirow{4}{*}{\rotatebox[origin=c]{90}{with MT}} & CE & 23.04 & 28.89 & 24.29 & 32.83 & 29.86 & 25.09 & 26.06 & 29.49 & 22.32 & 15.05 & 25.62 \\
             & CTC & 24.08 & 29.91 & 25.41 & 33.60 & 30.54 & 25.63 & 26.79 & 30.45 & \textbf{23.80} & 15.92 & 26.52 \\
             & CTC+CE & 24.28 & 30.21 & 25.01 & \textbf{34.01} & \textbf{30.55} & 25.49 & 26.94 & \textbf{30.56} & 23.76 & \textbf{16.23} & 26.57 \\
             & CTC+OT & \textbf{24.74} & \textbf{30.31} & \textbf{25.53} & 33.89 & 30.51 & \textbf{26.17} & \textbf{27.27} & 30.49 & 23.55 & 15.97 & \textbf{26.67} \\
             \bottomrule
        \end{tabular}
    }
    
\end{table*}

\subsection{Results without MT pre-training}
Results are shown in the upper halves of Tables~\ref{tab:covost_o2m_dev} and~\ref{tab:mustc_dev} for \covostt and \mustc, respectively.%

\paragraph{Bilingual} The results (2\textsuperscript{nd} and 3\textsuperscript{rd} columns of Table~\ref{tab:mustc_dev}, upper half) highlight the significant contribution of CTC in the final ST performance: all three methods using CTC in pre-training surpass the common method using CE (\numrange{0.89}{1.40} for En-De, and \numrange{1.35}{1.39} for En-Fr). Specifically, pre-training with CTC alone outperforms pre-training with CE by a large margin ($+0.89$ and $+1.39$, respectively, for En-De and En-Fr). Adding CE or OT on top of CTC is helpful for En-De, though it slightly hurts the performance on En-Fr.

\paragraph{Multilingual} The results (Tables~\ref{tab:covost_o2m_dev} and~\ref{tab:mustc_dev}, upper halves) show that CTC consistently outperforms CE on both datasets: the scores are higher across all language pairs by a large margin (on average, $+1.1$ and $+1.2$ on \mustc and \covostt, respectively). Adding CE on top of CTC slightly degrades the results ($26.47$ \vs $26.51$) while using OT marginally improves the performance ($26.60$ \vs $26.51$).

\subsection{Results with MT pre-training}

Results are shown in the lower halves of Tables~\ref{tab:covost_o2m_dev} and~\ref{tab:mustc_dev} for \covostt and \mustc, respectively.

\paragraph{Bilingual} The results (2\textsuperscript{nd} and 3\textsuperscript{rd} columns of Table~\ref{tab:mustc_dev}, lower half) show that CTC outperforms CE by around $+1.0$ BLEU point for both En-De and En-Fr. Compare with the without-MT-pre-training results (same columns but in the upper halves), we observe that MT pre-training are helpful for all methods. Especially, the gains for our proposed method are the highest ($0.83$/$0.48$ on En-De/En-Fr, compared to $0.35$/$0.41$, $0.50$/$0.04$, and $0.19$/$0.47$ for CE, CTC, and CTC+CE, respectively).

\paragraph{Multilingual} The results (Tables~\ref{tab:covost_o2m_dev} and~\ref{tab:mustc_dev}, lower halves) again show that CTC remains superior to CE. On the other hand, MT pre-training neither helps nor hurts the performance for \mustc, while the improvements are more noticeable on \covostt (\numrange{0.4}{0.7} on average).

\begin{table*}[!t]
	\addtolength{\tabcolsep}{-0.2em}
	\centering
    \caption{BLEU on \covostt test sets for En$\to$X. \textsuperscript{$\dagger$}results from baselines in~\cite{wang2020covost2}. ``CE'' is the same model as the corresponding ones in \covostt but with MT pre-training.
	\label{tab:covost_o2m_test}}
		\begin{tabular}{llccc ccc ccc ccc ccc a}
			\toprule
			& \textbf{Method} & \textbf{ar} & \textbf{ca} & \textbf{cy} & \textbf{de} & \textbf{et} & \textbf{fa} & \textbf{id} & \textbf{ja} & \textbf{lv} & \textbf{mn} & \textbf{sl} & \textbf{sv} & \textbf{ta} & \textbf{tr} & \textbf{zh} & \textbf{avg}\\
			\midrule

			\multirow{5}{*}{\rotatebox[origin=c]{90}{Medium}} & \covostt\textsuperscript{$\dagger$}& 11.2 & 21.6 & 22.9 & 15.9 & 12.8 & 13.8 & 19.7 & 31.5 & 12.4 & 9.2 & 15.2 & 21.5 & 10.6 & 9.8 & 29.3 & 17.2 \\ 
			
			& CE & 12.2 & 23.1 & 24.5 & 17.3 & 13.9 & 15.1 & 21.7 & 32.0 & 13.5 & 10.5 & 16.5 & 22.7 & 11.9 & 11.1 & 29.3 & 18.4 \\
			
			& CTC & 12.9 & 24.1 & 25.4 & 18.1 & 14.7 & 15.7 & 22.7 & 32.8 & 14.3 & 11.1 & 17.7 & 23.6 & 12.6 & 11.7 & 30.5 & 19.2 \\
			
			& CTC+CE & 13.1 & 24.3 & 25.6 & 18.2 & 14.8 & 15.9 & 22.8 & 32.8 & 14.4 & 11.1 & 17.7 & 24.0 & 12.9 & 11.8 & 30.3 & 19.3 \\
			
			& CTC+OT & \textbf{13.5} & \textbf{25.1} & \textbf{26.2} & \textbf{18.9} & \textbf{15.5} & \textbf{16.3} & \textbf{23.2} & \textbf{33.4} & \textbf{14.9} & \textbf{11.6} & \textbf{18.3} & \textbf{24.5} & \textbf{13.3} & \textbf{12.1} & \textbf{31.0} & \textbf{19.9} \\		
			\midrule
			\multirow{4}{*}{\rotatebox[origin=c]{90}{Large}} & \covostt\textsuperscript{$\dagger$}& 13.9 & 23.6 & 25.1 & 18.4 & 15.1 & 15.5 & 22.0 & 33.0 & 15.2 & 11.0 & 18.3 & 24.1 & 12.8 & 11.7 & 31.3 & 19.4  \\ 
			
			& CE & 13.5 & 23.8 & 25.2 & 18.3 & 14.9 & 15.7 & 22.6 & 32.8 & 14.6 & 11.1 & 17.8 & 23.8 & 12.6 & 11.6 & 30.1 & 19.2 \\
			
			& CTC & 13.8 & 24.5 & 25.9 & 18.8 & 15.4 & 16.1 & 23.0 & 33.6 & 15.2 & 11.5 & 18.6 & 24.4 & 13.2 & 11.9 & 31.1 & 19.8 \\ %
			
			& CTC+OT & \textbf{15.3} & \textbf{26.5} & \textbf{28.1} & \textbf{20.6} & \textbf{17.1} & \textbf{17.5} & \textbf{25.1} & \textbf{35.7} & \textbf{16.6} & \textbf{12.8} & \textbf{20.6} & \textbf{26.5} & \textbf{14.8} & \textbf{13.5} & \textbf{32.4} & \textbf{21.5} \\
	
			\bottomrule
		\end{tabular}
\end{table*}

\begin{table*}[!htb]
	\centering
     \caption{BLEU on \covostt test sets for X$\to$En. \textsuperscript{$\dagger$} results from baselines in~\cite{wang2020covost2}. ``CE'' is the same model as the corresponding ones in \covostt but with MT pre-training.
    	\label{tab:covost_m2o_test}}
	\begin{tabular}{ll cccc a}
		\toprule
		& \textbf{Method} & \textbf{fr} & \textbf{de} & \textbf{es} & \textbf{ca} & \textbf{avg} \\
		\midrule
		
		\multirow{5}{*}{\rotatebox[origin=c]{90}{Medium}} & \covostt\textsuperscript{$\dagger$} & 27.0 & 18.9 & 28.0 & 23.9 & 24.5  \\ 
		
		& CE &  27.9 & 19.3 & 28.3 & 22.9 & 24.6 \\
		
		& CTC & 27.7 & 19.7 & 28.2 & 23.2 & 24.7 \\
		
		& CTC+CE & 27.4 & 19.4 & 28.2 & 22.9 & 24.5 \\
		
		& CTC+OT & \textbf{28.4} & \textbf{20.4} &  \textbf{29.2} & \textbf{24.1} & \textbf{25.5}\\		
		\bottomrule
	\end{tabular}
	
\end{table*}

\begin{table*}[t]
\addtolength{\tabcolsep}{-0.2em}
\centering
\caption{\label{tab:mustc_test_results_full}BLEU on \mustc test sets (\texttt{tst-COMMON}), comparing with state-of-the-art methods. ``Multi'' means multilingual models. Note that ``CE pre-training'' (8\textsuperscript{th} row from the bottom) is the same as \textsc{fairseq S2T}~\cite{wang2020fairseq} (1\textsuperscript{st} row) but with MT pre-training, which turns out to be unhelpful.
}
\resizebox{\linewidth}{!}{
	 \begin{tabular}{lrc cc cccc cccc  a}
		\toprule
		\multicolumn{2}{l}{\multirow{2}{*}{\textbf{Method}}} & \multirow{2}{*}{\textbf{Multi}} & \multicolumn{2}{c}{\textbf{External Data}} & \multicolumn{9}{c}{\textbf{BLEU}} \\
		\cmidrule(lr){4-5}\cmidrule(l){6-14}
		& & & \textit{Unlabeled} & \textit{Labeled} & \textbf{de} & \textbf{es} & \textbf{fr} & \textbf{it} & \textbf{nl} & \textbf{pt} & \textbf{ro} & \textbf{ru} & \textbf{avg} \\ 
		
		\midrule
		
		\multicolumn{2}{l}{\textsc{fairseq S2T}~\cite{wang2020fairseq}}& \checkmark& - & - & 
		24.5 & 28.2 & 34.9 & 24.6 & 28.6 & 31.1 & 23.8 & 16.0 & 26.5 \\
		
		\multicolumn{2}{l}{ESPnet-ST~\cite{inaguma2020espnet}} & \checkmark  & -  & - & 
		22.9 & 28.0 & 32.7 & 23.8 & 27.4 & 28.0 & 21.9 & 15.8 & 25.1  \\
		
		\multicolumn{2}{l}{NeurST~\cite{zhao2021neurst} }& \checkmark & - & - & 
		22.8 & 27.4 & 33.3 & 22.9 & 27.2 & 28.7 & 22.2 & 15.1 & 24.9 \\
		
		\multicolumn{2}{l}{Dual-decoder~\cite{le2020dual} }& \checkmark  & - & - & 
		23.6 &28.1 & 33.5 & 24.2 & 27.6 & 30.0 & 22.9 & 15.2 & 25.6 \\
		
		\multicolumn{2}{l}{Adapters~\cite{le2021lightweight}} & \checkmark & - & -    & 
		24.7 & 28.7 & 35.0 & 25.0 &  28.8 & 31.1 & 23.8 & 16.4 & 26.6 \\

		\multicolumn{2}{l}{AFS~\cite{zhang2020adaptive}} & - & - & - & 
		22.4 & 26.9 & 31.6 & 23.0 & 24.9 &26.3 & 21.0 & 14.7 & 23.9 \\
		
		\multicolumn{2}{l}{Speechformer~\cite{papi2021speechformer} }& - & - & - & 
		23.6 & 28.5 & - & -&  27.7 & - & - & - & - \\
		
		\multicolumn{2}{l}{Mutual-learning~\cite{zhao2021mutual}} & - & - & - &
		- & 28.7 & 36.3 & -& - &-&-& - & - \\
		
		\multicolumn{2}{l}{BiKD~\citep{inaguma2021source} }& - & - & - & 
		25.3 & - & 35.3 & - & - & - & - & -  & - \\

        \multicolumn{2}{l}{TDA~\cite{du2022regularizing}}& - & - & - & 25.4 & 29.6 & 36.1 & 25.1 & 29.6 & 31.1 & 23.9 & 16.4 & 27.2 \\
		
		\multicolumn{2}{l}{Adversarial~\cite{alinejad2020effectively}} & - & - & \checkmark & 20.2 & - & 17.0 & - & - & - & - & - & - \\
		
		\multicolumn{2}{l}{MTL~\cite{tang2021general}}  & - & - & \checkmark  & 
		23.9 & 28.6 & 33.1 & - & - & - & - & - & -\\
		
		\multicolumn{2}{l}{Joint Speech Text~\cite{tang2021improving}} & - & - & \checkmark &
		26.8 & 31.0 & 37.4 &- &-&-&-& - & - \\

		\multicolumn{2}{l}{SATE~\cite{xu2021stacked} } & - & -  & \checkmark  & 
		28.1  & - & -&-&-&-& -  & - & -\\ 
		
		\multicolumn{2}{l}{TaskAware~\cite{indurthi2021task} }& - & - & \checkmark  &
		\textbf{28.9}& - &- &-&-&-& - & - & - \\
		
		\multicolumn{2}{l}{Self-training~\citep{pino2020self} }& - & \checkmark & \checkmark &
		25.2 & - & 34.5 & - & - & - & - & -  & -\\

		\multicolumn{2}{l}{FAT-ST (Big)~\citep{zheng2021fused} } & -  & \checkmark  & \checkmark   & 
		25.5 & 30.8 & - & - & 30.1 & - & - & - & - \\

		\multicolumn{2}{l}{XSTNet~\cite{ye2021end}} & - &  \checkmark  & \checkmark & 27.1 & 30.8 & 38.0 & 26.4 & 31.2 & 32.4 & 25.7 & 18.5 & 28.8 \\
		
		\multicolumn{2}{l}{Chimera~\cite{han2021learning}}  & - & \checkmark & \checkmark & 27.1 & 30.6 & 35.6 & 25.0 & 29.2 & 30.2 & 24.0 & 17.4  & 27.4 \\

		\multicolumn{2}{l}{STEMM~\cite{fang2022stemm}} & - & \checkmark & \checkmark & 			28.7 & 31.0 & 37.4 & 25.8 & 30.5 & 31.7 & 24.5 & 17.8 & 28.4 \\
		
		\multicolumn{2}{l}{ConST~\cite{ye2022cross} } & - & \checkmark &  \checkmark  &
		28.3  & 32.0 & 38.3 & 27.2 & \textbf{31.7} & 33.1 & 25.6 & \textbf{18.9} & 29.4 \\ 
		
		\multicolumn{2}{l}{STPT~\cite{tang2022unified}} & - & \checkmark & \checkmark & 
		- & \textbf{33.1} & \textbf{39.7} &-&-&-&-& - & -\\
		
		\midrule
		
		CE pre-training & \multirow{4}{*}{\textsc{medium}} & \checkmark & - & - & 24.6 & 28.7 & 34.9 & 24.6 & 28.4 & 30.7 & 23.7 & 15.9 & 26.4 \\
		CTC pre-training&  & \checkmark & - & - & 25.9 & 29.7 & 36.6 & 25.6 & 29.6 & 32.0 & 24.6 & 16.7 & 27.6 \\
		CTC+CE pre-training & & \checkmark & - & - & 25.6 & 29.5 & 36.4 & 25.2 & 29.5 & 31.6 & 24.5 & 16.5 & 27.4 \\
		Siamese-PT (this work)& 
		& \checkmark & - & - & 26.2 & 29.8 & 36.9 & 25.9 & 29.8 & 32.1 & 24.8 & 16.8 & 27.8 \\
		\midrule

		CE pre-training&  \multirow{4}{*}{\textsc{large}}& \checkmark & - & - & 26.9 & 30.8 & 37.7 & 26.7 & 30.8 & 33.3 & 26.2 & 17.9 & 28.8 \\
		
		CTC pre-training& & \checkmark & - & - & 27.6 & 31.4 & 38.2 & 27.2 & 31.1 & 33.6 & 26.4 & 18.4 & 29.2 \\
		
		CTC+CE pre-training& & \checkmark & - & - & 27.2 & 31.2 & 38.0 & 27.0 & 31.5 & 33.7 & 26.2 & 18.3 & 29.1 \\
		
		Siamese-PT (this work)& & \checkmark & - & - & 27.9 & 31.8 & 39.2 & \textbf{27.7}& \textbf{31.7} & \textbf{34.2} & \textbf{27.0} & 18.5 & \textbf{29.8} \\
		
		\bottomrule
	\end{tabular}
}

\end{table*}

\end{document}